\documentclass[journal]{IEEEtran}
\usepackage{blindtext}
\usepackage{graphicx}
\usepackage{amssymb}
\usepackage{graphicx}
\usepackage{bbm}
\usepackage{mathtools}
\usepackage{mathrsfs}

\usepackage{multirow}
\usepackage{algorithm}
\usepackage{algorithmic}
\usepackage{bbding}
\usepackage[table,xcdraw]{xcolor}
\usepackage{epsfig}

\usepackage{amsmath}
\usepackage{subfigure}
\usepackage{times}
\usepackage{epsfig}
\usepackage{graphicx}
\usepackage{amsmath}
\usepackage{amssymb}
\usepackage{bm}
\usepackage{amsfonts,amssymb}
\usepackage{booktabs}
\usepackage{float}

\usepackage{subfigure}
\usepackage{makecell}
\usepackage{multirow}
\usepackage{hyperref}
\newcommand{\citet}[1]{\citeauthor{#1} \shortcite{#1}}

\makeatletter
\DeclareRobustCommand\onedot{\futurelet\@let@token\@onedot}
\def\@onedot{\ifx\@let@token.\else.\null\fi\xspace}

\makeatother

\ifCLASSINFOpdf
\else
\fi
\begin{document}
%
\title{COLA-Net: Collaborative Attention Network for Image Restoration}
\author{Chong~Mou, Jian~Zhang, Xiaopeng~Fan, Hangfan~Liu, Ronggang~Wang 

\thanks{
C.~Mou, J.~Zhang, and R.~Wang are with the School of Electronic and Computer Engineering, Peking University Shenzhen Graduate School, Shenzhen, China. (e-mail: eechongm@stu.pku.edu.cn; zhangjian.sz@pku.edu.cn; rgwang@pkusz.edu.cn). 

X.~Fan is with the School of Computer Science and Technology, Harbin Institute of Technology, China (e-mail: fxp@hit.edu.cn). 

H.~Liu is with Center for Biomedical Image Computing \& Analytics, University of Pennsylvania, USA (e-mail: hfliu@upenn.edu).  

}
}

\markboth{2021}%
{Shell \MakeLowercase{\textit{et al.}}: Bare Demo of IEEEtran.cls for IEEE Journals}

\maketitle

\begin{abstract}
Local and non-local attention-based methods have been well studied in various image restoration tasks while leading to promising performance. However, most of the existing methods solely focus on one type of attention mechanism (local or non-local). Furthermore, by exploiting the self-similarity of natural images, existing pixel-wise non-local attention operations tend to give rise to deviations in the process of characterizing long-range dependence due to image degeneration. To overcome these problems, in this paper we propose a novel collaborative attention network (COLA-Net) for image restoration, as the first attempt to combine local and non-local attention mechanisms to restore image content in the areas with complex textures and with highly repetitive details respectively. In addition, an effective and robust patch-wise non-local attention model is developed to capture long-range feature correspondences through 3D patches. Extensive experiments on synthetic image denoising, real image denoising and compression artifact reduction tasks demonstrate that our proposed COLA-Net is able to achieve state-of-the-art performance in both peak signal-to-noise ratio and visual perception, while maintaining an attractive computational complexity. The source code is available on \href{https://github.com/MC-E/COLA-Net}{https://github.com/MC-E/COLA-Net}.
\end{abstract}

\begin{IEEEkeywords}
Image restoration, deep neural network, image denoising, non-local attention, feature fusion.
\end{IEEEkeywords}

%
\IEEEpeerreviewmaketitle

\section{Introduction}

\IEEEPARstart{I}{mage} restoration aims to recover the underlying high-quality image $\mathbf{x}$ from its degraded measurement $\mathbf{y} = \mathbf{A}\mathbf{x} + \mathbf{n}$, where $\mathbf{A}$ is a linear degradation matrix, and $\mathbf{n}$ represents additive noise. It is typically an ill-posed problem due to the irreversible degradation process \cite{zhang2014csvt,zhang2014imageSP,zhao2018cream}. Since deep learning methods have been successfully applied in various computer vision tasks, many deep-learning-based methods \cite{dncnn,ffdnet,ircnn,memnet} have been proposed to solve this ill-posed problem. Most of them focus on local processing by combining convolutional layers and element-wise operations. To aggregate more useful information from a large receptive field, some strategies such as hourglass-shaped architecture \cite{cbdnet,vdnet,unprocessingnet,aindnet}, dilated convolutions \cite{dilated,dilated2,ircnn} or stacking more convolutional layers \cite{rnan,memnet,red,sb2} are applied in image restoration tasks. Furthermore, inspired by some traditional excellent image restoration algorithms such as BM3D \cite{BM3D}, group sparse representation \cite{GSR2014}, and non-local means \cite{nonlocalmean}, which take advantage of self-similarity within the whole image to recover a local image content. Some recent works tried to implement this idea via deep networks. One way is to establish the long-range dependence based on pixels, \textit{e.g.}, NLRN \cite{nlrn} and RNAN \cite{rnan}, which applied non-local neural networks \cite{nonlocalnet} in image restoration tasks. However, due to the fact that pixels are usually noisy in image restoration tasks, 
establishing relationships between pixels is prone to be biased and unreliable, especially when images are heavily corrupted. In addition to non-local attention mechanism, local attention mechanism is also an important strategy in computer vision community and has been studied in many image restoration tasks \cite{ridnet, dilated, mirnet}. However, the main drawback of these methods is that the receptive field during image restoration is relatively small.

Through the above analysis, both local and non-local attentions have their unique drawbacks, but they can complement each other in many aspects. For instance, when an image contains sufficient repetitive details, non-local operations will be more useful. But when an image has a lot of complex textures, directly applying non-local operations will cause over-smooth artifacts, while local operations may be an appropriate choice. How to make a trade-off between local and non-local operations in image restoration has not been explored.

To address the above issues, we present the first attempt to exploit both local attention and non-local attention to restore image content in areas with complex textures and highly repetitive details, respectively. It is important to note that this combination is learnable and self-adaptive. Part of our previous work has been reported in \cite{SAT-Net}. To be concrete, for local attention operation, we apply local channel-wise attention on different scales to enlarge the size of receptive field of local operation, while for non-local attention operation, we develop a novel and robust patch-wise non-local attention model for constructing long-range dependence between image patches to restore every patch by aggregating useful information (self-similarity) from the whole image. \textbf{The main contributions of this paper are summarized as follows:}
\begin{itemize}
\item We propose a novel \textbf{COL}laborative \textbf{A}ttention \textbf{Net}work, dubbed \textbf{COLA-Net}, which incorporates both local and non-local attention mechanisms into deep networks for image restoration tasks. To the best of our knowledge, COLA-Net is the first attempt to combine local and non-local operations to restore complex textures and repetitive details distinguishingly.
\item We propose an effective patch-wise non-local attention model to establish a more reliable long-range dependence during image restoration.
\item We carry out extensive experiments on three typical image restoration tasks, i.e., synthetic image denoising, real image denoising, and compression artifact reduction, showing that our proposed COLA-Net achieves state-of-the-art results while maintaining an attractive computational complexity.
\end{itemize}

\section{Related Work}

In what follows, we give a brief review of both local and non-local attention mechanisms for image restoration and focus on the specific methods most relevant to our own.

\textbf{Non-local Attention.} Self-similarity is an important prior to image restoration especially for image denoising, and this prior has been widely used in several traditional non-local image restoration methods \cite{nonlocalmean,BM3D,GSR2014,Liu_2015_CVPR,zhao2016reducing,CONCOLOR}. In general, non-local attention models take a degraded input $\mathbf{y} = \{\mathbf{y}_{i} | i \in \mathbb{I}\}$, where $\mathbb{I}$ denotes the set of indices of pixels/patches in the whole image, and $\mathbf{y}_{i}$ stands for the $i$-th pixel or the patch centered on the $i$-th pixel. Each corrupted element $\mathbf{y}_{i}$ can be restored by a set of similar items $\mathbf{y}_{j}$ from a search region $\mathbb{Q} \subset \mathbb{I}$. Thus, non-local operations can be formally defined as:
\begin{equation}
    \hat{\mathbf{x}}_{i} = \frac{1}{z_{i}}\sum_{{j}\in \mathbb{Q}}\phi(\mathbf{y}_{i},\mathbf{y}_{j})G(\mathbf{y}_{j}), \forall i,
\label{eq1}
\end{equation}
where $\hat{\mathbf{x}}_{i}$ and $z_{i}$ represent the estimated clean value and the normalizing constant calculated by $z_{i}=\sum_{{j}\in \mathbb{Q}}\phi(\mathbf{y}_{i},\mathbf{y}_{j})$. The function $\phi$ can compute the similarity between query items $\mathbf{y}_{i}$ and key items $\mathbf{y}_{j}$. $G$ is an embedding function to transform $\mathbf{y}_{j}$ to another representation. 

The seminal work non-local means \cite{nonlocalmean} searched similar patches within a local region and averaged central pixels weighted by similarity. This method applied weighted Euclidean distance with Gaussian kernel to compute the similarity between two patches, which can be formulated as:
\begin{equation}
       \phi(\mathbf{y}_{i},\mathbf{y}_{j})=e^{-\frac{\left\|\mathbf{y}_{i}-\mathbf{y}_{j}\right\|_{2,\alpha}^{2}}{h^{2}}},
\end{equation}
where $\alpha$ and $h$ refer to the standard deviation of Gaussian kernel and the filter factor, respectively. Identity mapping: $G(\mathbf{y}_{j})=\mathbf{y}_{j}$ is directly used as the embedding function in this model. Rather than simply averaging similar pixels, the popular method BM3D \cite{BM3D} generated a stack of 3D matching patches and utilized a 3D filter to restore corrupted patches. 

Recently, deep neural networks (DNN) have been prevalent in the community of computer vision, and non-local neural network \cite{nonlocalnet} has been proposed for high-level vision tasks such as object detection and classification. In \cite{nonlocalnet}, embedding function $G$ is a convolutional layer that can be viewed as a linear embedding function, and dot product, i.e., $\phi(\mathbf{y}_i,\mathbf{y}_j)=\mathbf{y}_i^T\mathbf{y}_{j}$ is applied to compute the similarity between query and keys. Based on the success of non-local neural networks, NLRN \cite{nlrn} and RNAN \cite{rnan} were proposed for image restoration tasks. In NLRN \cite{nlrn}, the distance matrice of non-local layers is shared to enable the feature correlation to be propagated along with adjacent recurrent states. In RNAN \cite{rnan}, a residual non-local attention learning was proposed to train very deep networks by preserving more low-level features. However, pixel-wise non-local attention is unreliable due to the image degeneration. Another way is to establish the long-range dependence based on patch matching and restore local patch or central pixel through similar patches in other places, \textit{e.g.}, \cite{nlnet,nldiff,unlnet}. Nevertheless, the patching matching step is isolated from the training process. Therefore it can not be jointly trained with image restoration networks. In \cite{sb1}, a light-weight model was proposed to utilize non-local attention and sparsity principles among feature patches for image restoration. In addition, N3Net \cite{n3net} and \cite{irpnet} proposed learnable patch matching. Since they can only match a small number of blocks within a limited area, these two methods are not efficient enough as pixel matching mechanisms. 


\textbf{Local Attention.} Compared with non-local attention, local attention is originally designed for high-level vision tasks \cite{cbam,SENet,SKNet}. RCAN \cite{rcan} proposed a very deep residual channel attention networks for highly accurate image super-resolution. RIDNet \cite{ridnet} was the first attempt to combine local attention in image denoising tasks. Later on, MIRNet \cite{mirnet} expanded local attention to a wider range of image restoration tasks. To enlarging the receptive field, ADNet \cite{dilated2} applied dilated convolution for denoising tasks. However, compared with non-local methods, the main drawback of local attention mechanism during image restoration is the relatively small receptive field.

\textbf{Image Restoration Architectures.} Stacking convolutional layers is the most well-known CNN-based strategy for image restoration. Dong \textit{et al}. proposed ARCNN \cite{arcnn} for image compression artifact reduction with several stacked convolutional layers. Zhang \textit{et al}. proposed DnCNN \cite{dncnn} for image restoration with the help of residual learning and batch normalization. Based on DnCNN, Zhang \textit{et al}. proposed FFDNet \cite{ffdnet} for blind image denoising, which improves the generalization of DNN-based image restoration methods. Zhang \textit{et al.} further proposed IRCNN \cite{ircnn}, which applied dilated convolution in image restoration tasks. Recently, great progress has been made in the image restoration community, and diverse novel models have been proposed. Tai \textit{et al}. proposed MemNet \cite{memnet} to apply dense connection in convolutional layers, and He \textit{et al.} \cite{mrfn} applied this idea to a light-weight design. Tian et al. \cite{c2f} proposed a coarse-to-fine architecture to perform image super-resolution. Guo \textit{et al}. proposed CBDNet \cite{cbdnet}, making a sufficient improvement in dealing with real-noise image corruption problems, and its hourglass-shaped architecture is applied in many image restoration works \cite{vdnet,unprocessingnet,aindnet} afterward.

\begin{figure*}[t]
  \centering
  \includegraphics[width=.95\linewidth,height=6.3cm]{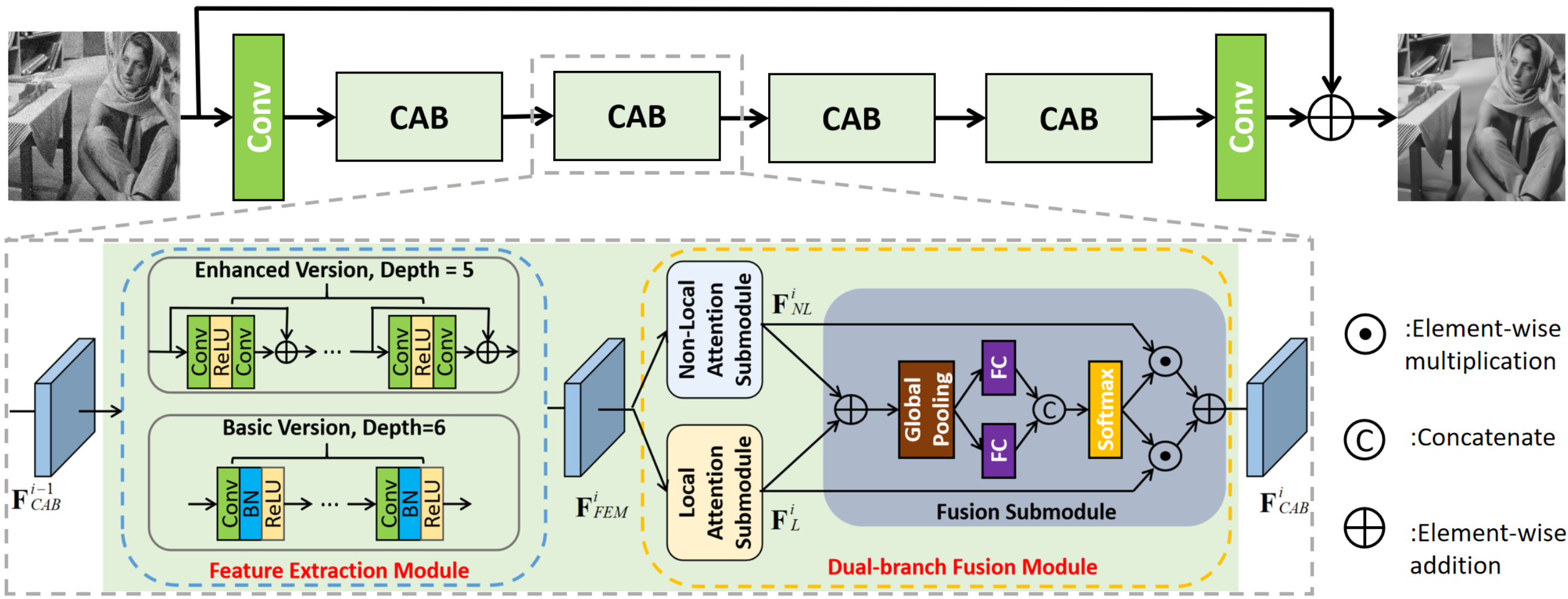}
  \caption{Illustration of the network architecture of our proposed COLA-Net. COLA-Net is mainly cascaded by several collaborative attention blocks (CAB). The second row further shows the detailed architecture of each CAB.}
\label{im_network} 
\end{figure*}

\section{Collaborative Attention Network (COLA-Net)}
In this section, we will elaborate our proposed COLlaborative Attention Network (COLA-Net), which is able to adjust the contribution of local operation and non-local operation adaptively in the process of image restoration.

\subsection{Framework}
An important concept presented in our work is that local attention operation and non-local attention operation are not all-powerful to restore any kind of image content. Using these two operations without distinction will produce passive effects. To rectify this weakness, we propose a novel collaborative attention network (COLA-Net) by combining local attention submodule and non-local attention submodule to restore complex textures and repetitive details, respectively. Note that this combination is not a simple series connection or parallel connection but is self-adaptive and learnable. 

As illustrated in Fig.~\ref{im_network}, our proposed COLA-Net is mainly composed of several collaborative attention blocks (CAB) (four by default) that are cascaded together. Let's denote $\mathbf{I}_{LQ}$ and $\mathbf{I}_{REC}$ as the low-quality (LQ) input and the reconstruction output of COLA-Net. We first use one convolutional layer to extract the shallow features $\mathbf{F}_{S}$ from $\mathbf{I}_{LQ}$:
\begin{equation}
    \mathbf{F}_{S} = \mathcal{H}_{SF}(\mathbf{I}_{LQ}),
\end{equation}
where $\mathcal{H}_{SF}(\cdot)$ denotes the shallow feature extraction. $\mathbf{F}_{S}$ is then used for feature restoration with collaborative attention blocks (CAB). So we can further have:
\begin{equation}
\mathbf{F}_{CAB}^{i} = \mathcal{H}_{CAB}^{i}(...\mathcal{H}_{CAB}^{1}(\mathbf{F}_{S})),
\end{equation}
where $\mathcal{H}_{CAB}^{i}(\cdot)$ and $\mathbf{F}_{CAB}^{i}$ denote the function of the $i$-th CAB and its corresponding restoration result, $i=1,2,3,4$. The architecture of CAB is shown in the second row of Fig.~\ref{im_network}, which is composed of two parts, i.e., feature exaction module (FEM) and dual-branch fusion module (DFM). In this paper, we provide two versions of FEM. In consideration of the computational complexity, we develop a basic version of FEM that applies a DnCNN \cite{dncnn} architecture, containing six convolutional blocks. Each block consists of a 3 $\times$ 3 convolutional layer with 64 filters, a batch normalization layer, and a ReLU activation function. Considering the depth of neural networks is important for image restoration, an enhanced version of FEM in the residual domain is derived to further improve restoration performance with moderate growth of parameters. The COLA-Net equipped with the basic FEM is named COLA-B, and the COLA-Net equipped with the enhanced FEM is named COLA-E. All convolutional layers in FEM have 64 filters and $3\times 3$ kernel size. In the experiment, we set the depth of the basic FEM to six and the enhanced FEM to five. The output of FEM in the $i$-th CAB is denoted by $\mathbf{F}_{FEM}^{i}$. DFM is the core of COLA-Net, which performs local attention operation and non-local attention operation in two parallel branches and fuses them adaptively. More details of DFM will be given in the next subsection. 

At the end of the network, we apply one convolutional layer to transform the output of the last CAB from feature domain to image domain and adopt residual learning to facilitate network training. Thus, we can get the output $\mathbf{I}_{REC}$ as:
\begin{equation}
    \mathbf{I}_{REC} = \mathbf{I}_{LQ} + \mathcal{H}_{DF}(\mathbf{F}_{CAB}^{4}) = \mathcal{H}_{COLA}(\mathbf{I}_{LQ}),
\end{equation}
where $\mathcal{H}_{DF}(\cdot)$ and $\mathcal{H}_{COLA}(\cdot)$ represent the functions of the last convolutional layer and the whole COLA-Net, respectively. 

To show the effectiveness of COLA-Net, we choose the commonly used $L_{2}$ loss as the objective function. Given a training set $\{\mathbf{I}_{LQ}^{b},\mathbf{I}_{HQ}^{b}\}_{b=1}^{B}$ that contains $B$ corrupted low-quality (LQ) inputs and their high-quality (HQ) labels. The goal of training can be defined as:
\begin{equation}
    L(\mathbf{\Theta} )=\frac{1}{B}\sum_{b=1}^{B}\left\|\mathbf{I}_{HQ}^{b}-\mathcal{H}_{COLA}(\mathbf{I}_{LQ}^{b})\right\|_{2}^{2},
\end{equation}
where $\mathbf{\Theta}$ refers to the learnable parameters of COLA-Net.

\begin{figure*}[t]
\centering
  \includegraphics[width=0.85\textwidth]{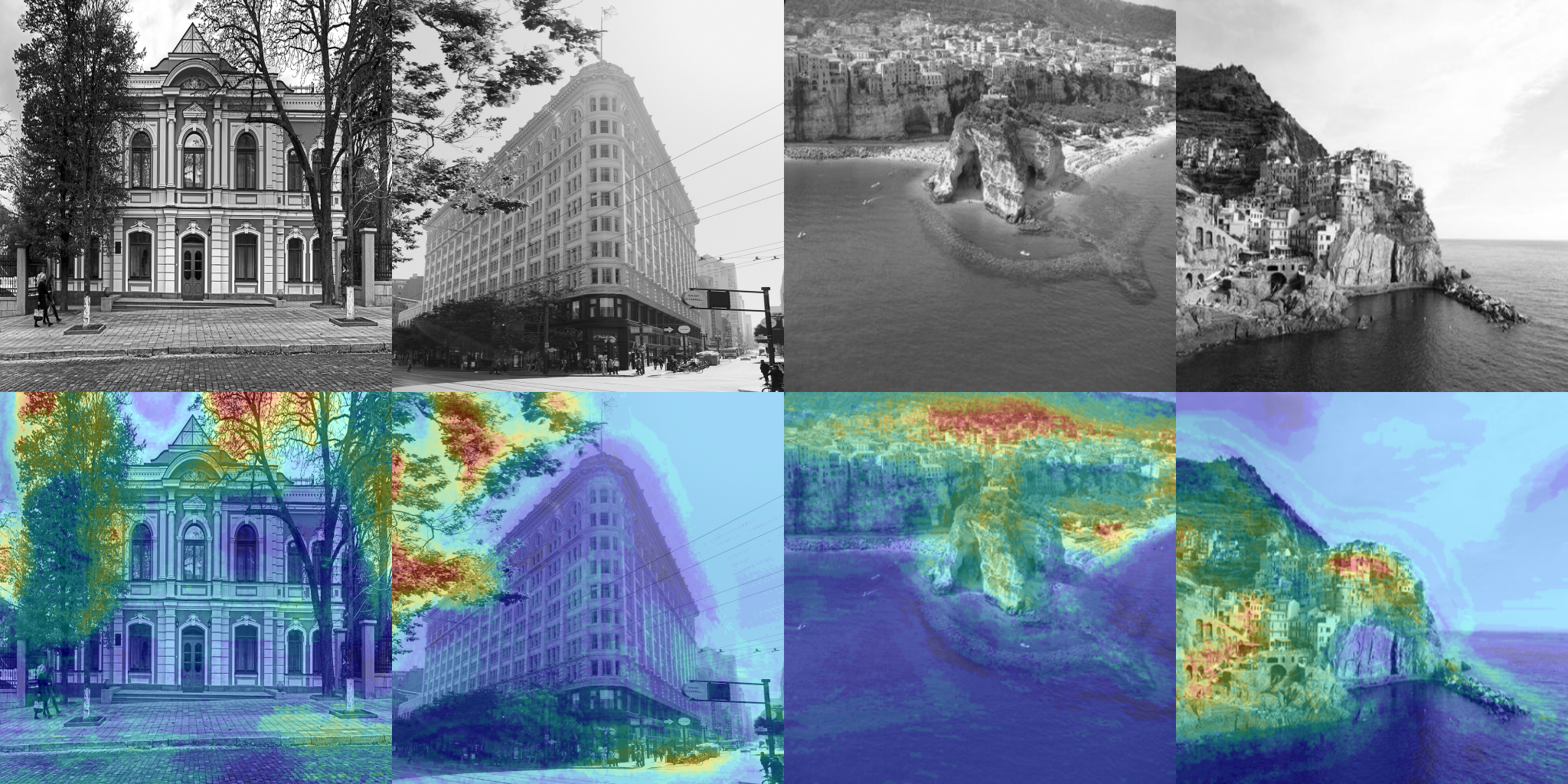}
\caption{Visualization of attention dependence during image restoration. The first row represents the original images (HQ) and the second row represents attention dependence during image restoration. For illustration purposes, the heat-maps directly overlap with original images in the second row. The deeper the color the more local operations are demanded, on the contrary, more non-local operations are demanded.}
\label{im_atdp}
\end{figure*}

\subsection{Dual-branch Fusion Module}
Our proposed dual-branch fusion module (DFM), labeled with a yellow box in Fig.~\ref{im_network}, comprises three parts: local attention submodule, non-local attention submodule, and fusion submodule. The role of DFM is to restore complex textures through local operation, recover repetitive details based on self-similarity, and fuse dual-branch restoration results in an adaptive way.

As illustrated in Fig.~\ref{im_network}, given $\mathbf{F}_{FEM}^{i}$, the local attention submodule and non-local attention submodule generate their outputs $\mathbf{F}^{i}_{L}$ and $\mathbf{F}^{i}_{NL}$ in a paralleled manner: 
\begin{equation}
    \begin{cases}
\mathbf{F}^{i}_{L}=\mathcal{H}_{L}(\mathbf{F}_{FEM}^{i})\\
\mathbf{F}^{i}_{NL}=\mathcal{H}_{NL}(\mathbf{F}_{FEM}^{i}),
\end{cases}
\label{eq_branch}
\end{equation}
where $\mathcal{H}_{L}(\cdot)$ and $\mathcal{H}_{NL}(\cdot)$ refer to the functions of the local and non-local attention submodules. The architecture of these two submodules will be elaborated in the following subsections. 

After getting $\mathbf{F}^{i}_{NL}$ and $\mathbf{F}^{i}_{L}$, and motivated by \cite{SKNet}, the final output $\mathbf{F}^{i}_{CAB}$ is obtained by the fusion submodule in the following three steps:
\begin{equation}
    \begin{cases}
 & \mathbf{v}= GlobalPooling(\mathbf{F}_{NL}^{i}+\mathbf{F}_{L}^{i})\\ 
 & \mathbf{w}_{NL},\mathbf{w}_{L}=softmax([\mathcal{H}_{FC}^{1}(\mathbf{v}),\mathcal{H}_{FC}^{2}(\mathbf{v})]) \\ 
 & \mathbf{F}^{i}_{CAB}=\mathbf{F}_{NL}^{i}\cdot \mathbf{w}_{NL}+\mathbf{F}_{L}^{i}\cdot \mathbf{w}_{L},
\end{cases}
\label{eq_merge}
\end{equation}
where $\mathcal{H}_{FC}^{1}(\cdot)$ and $\mathcal{H}_{FC}^{2}(\cdot)$ refer to two independent fully-connected layers. 

To be concrete, as shown in Eq. \ref{eq_merge}, we first merge the restoration results from dual branches via an element-wise addition. Then, we apply a global pooling to produce a global feature vector $\mathbf{v}\in \mathbb{R}^{C}$, which is used as the guidance of adaptive and accurate selection between local operation and non-local operation. Next, we apply two fully connected layers ($\mathcal{H}_{FC}^{1}$ and $\mathcal{H}_{FC}^{2}$) to generate two weight vectors ($\mathbf{w}_{NL}$ and $\mathbf{w}_{L}$) to perform channel-wise selection between two restoration results. Obviously, $\mathbf{w}_{NL}$ and $\mathbf{w}_{L}$ are content-aware, thus self-adaptive. In this way, local and non-local attention branches can restore image content selectively based on their characteristics.

\begin{figure}[t]
  \centering
  \includegraphics[width=0.98\linewidth,height=4cm]{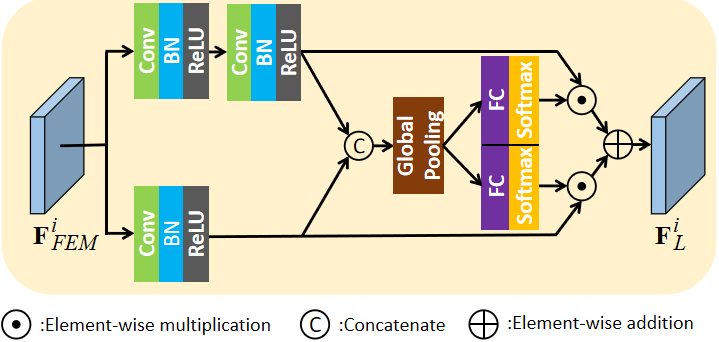}
  \caption{Illustration of our proposed local attention submodule.}
\label{im_attention} 
\end{figure}

\begin{figure*}[h]
\centering
\subfigure[]{
\includegraphics[width=0.24\linewidth]{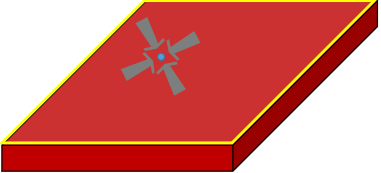}
}%
\subfigure[]{
\includegraphics[width=0.24\linewidth]{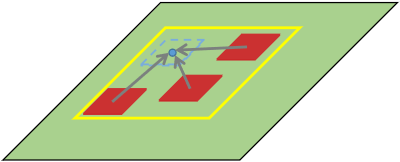}
}%
\subfigure[]{
\includegraphics[width=0.24\linewidth]{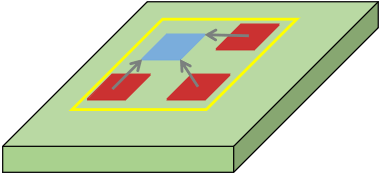}
}%
\subfigure[]{
\includegraphics[width=0.24\linewidth]{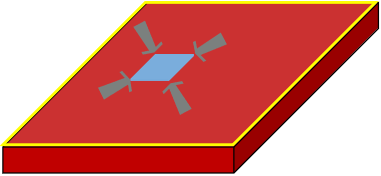}
}%
\centering
\caption{Visual comparison of various non-local operations. The blue parts and the red parts in images represent restored image content and corresponding long-range dependent items. The yellow boundary refers to the search region. (a) Common non-local neural networks \cite{nonlocalnet} applies pixel matching and pixel updating within the whole image. (b) Classic patch-wise non-local operation takes non-local means \cite{nonlocalmean} for example, applying patch matching and pixel updating within a limited region. (c) Learnable patch-wise non-local operation takes N3Net \cite{n3net} for example, applying learnable patch matching and patch updating with a limited region and limited matching items. (d) Our proposed patch-wise non-local model applies learnable patch matching and patch updating within the whole image.}
\label{im_nl_cp} 
\end{figure*}

\begin{figure*}[t]
  \centering
  \includegraphics[width=0.95\linewidth]{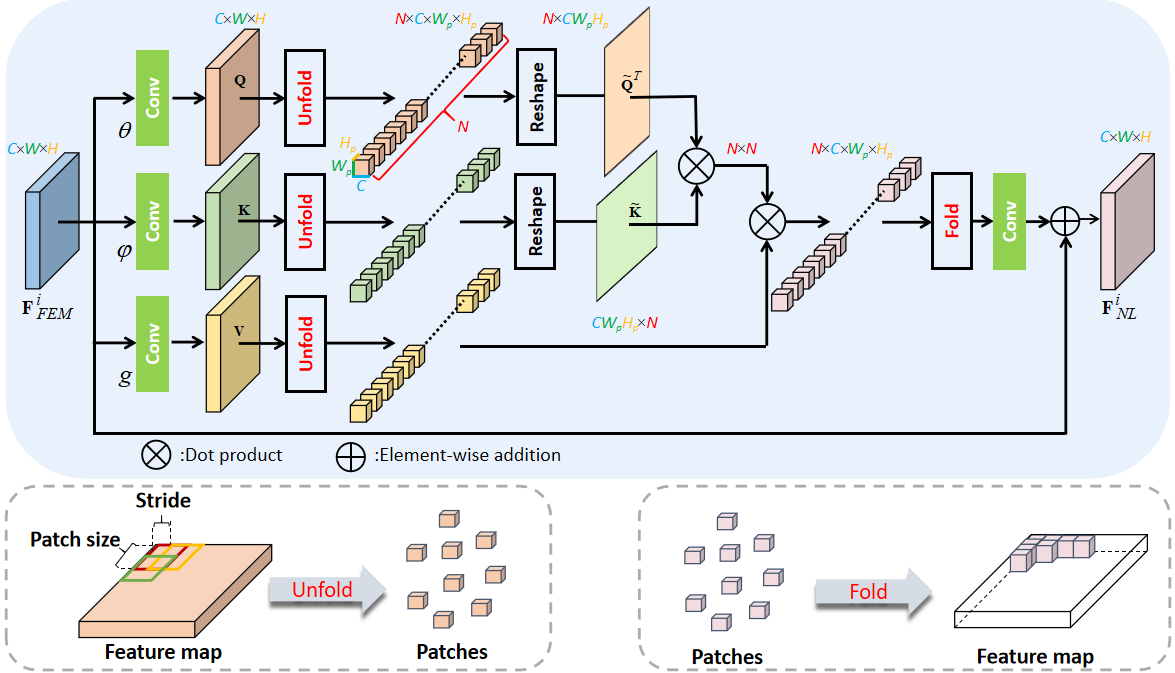}
  \caption{Illustration of our proposed non-local attention submodule. In particular, a novel effective and robust patch-wise non-local attention model is developed. The \textbf{unfold} operation is to extract sliding local patches from a batched input feature map, while the \textbf{fold} operation is to combine an array of sliding local patches into a large feature map.}
\label{im_nonlocal} 
\end{figure*}

In order to further verify the necessity of local and non-local attention results, we make a visualization of the attention dependence during image restoration by heat-maps. The heat value $h$ in Fig.~\ref{im_atdp} is computed as follows:
\begin{equation}
h=\frac{1}{M}\sum_{m=1}^{M}\mathbf{a}_{m},
\end{equation}
where $\mathbf{a}_{m}=1$ for $\mathbf{w}_{L}^{m}>=\mathbf{w}_{NL}^{m}$ and $0$ otherwise, and $M$ is the length of dependence weights ($\mathbf{w}_{NL}$ or $\mathbf{w}_{L}$). Thus, $h$ can represent the preference or focus between local and non-local attention operations in the process of image restoration in specific places. In Fig.~\ref{im_atdp}, the deeper the color, the more local attention operations are demanded. On the contrary, more non-local attention operations are needed. For illustration purposes, we overlap the heat-maps on original images in the second row of Fig.~\ref{im_atdp}. We can find that the heated parts are mainly concentrated in areas with complex textures or without sufficiently repetitive details indicating that these parts demand more local attention operations to restore. Conversely, light-colored areas contain a large amount of repetitive image content. The results of this experiment are entirely consistent with our motivation. 

\subsection{Local Attention Submodule}
Inspired by \cite{SKNet,SENet}, we employ the channel-wise attention mechanism to design the local attention submodule to perform channel selection with multiple sizes of receptive field. The detailed architecture of our local attention submodule is shown in Fig.~\ref{im_attention}. In this submodule, we adopt two branches to carry different numbers of convolutional layers to generate feature maps with different sizes of receptive field. The channel-wise attention is independently performed on these two outputs, and the results are added together. The whole process of local attention submodule can be represented by $\mathbf{F}^{i}_{L}=\mathcal{H}_{L}(\mathbf{F}^{i}_{FEM})$.

\subsection{Non-local Attention Submodule}
In this part, we will introduce our proposed patch-wise non-local attention model. To make a clear distinction, we first conduct a simple comparison and analysis of typical non-local methods, including patch-wise non-local operations and non-local neural networks \cite{nonlocalnet} (pixel-wise). Then we will give the details of our proposed patch-wise non-local attention model.

\textbf{Comparison and Analysis.} Fig.~\ref{im_nl_cp} shows a visual comparison of various non-local operations. The blue parts and the red parts in Fig.~\ref{im_nl_cp} refer to the restored image content and its corresponding long-range dependent items, respectively, while the yellow boundary refers to the search region of pixels/patches matching. In Fig.~\ref{im_nl_cp}(a), the non-local neural networks \cite{nonlocalnet} constructed the long-range dependence between pixels, and it applied a learnable embedding function to make the matching process adaptive. However, the long-range dependence based on pixels is unreliable in image restoration tasks due to image corruption. In Fig.~\ref{im_nl_cp}(b), the classic non-local means \cite{nonlocalmean} applied patch matching and pixel updating (only applied central pixels) to restore corrupted image content. This type of traditional method suffers from the drawback that parameters are fixed. Fig.~\ref{im_nl_cp}(c) shows the operation of learnable patch-wise non-local methods, and we take N3Net \cite{n3net} for example. This kind of method can perform patch matching and updating adaptively, but they are not efficient enough. The illustration of our method is shown in Fig.~\ref{im_nl_cp}(d). In what follows, we will present our solution to make the patch-wise non-local operation more effective and efficient.

\begin{figure}[t]
\centering
\small
\begin{minipage}{0.04\linewidth}
\centering
\includegraphics[width=1\linewidth,height=2cm]{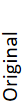}\\
\includegraphics[width=1\linewidth,height=2cm]{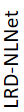}\\
\includegraphics[width=1\linewidth,height=2cm]{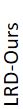}\\
\end{minipage}
\begin{minipage}{0.22\linewidth}
\centering
\includegraphics[width=1\linewidth]{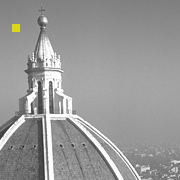}\\
\includegraphics[width=1\linewidth]{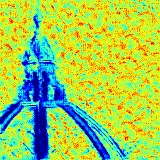}\\
\includegraphics[width=1\linewidth]{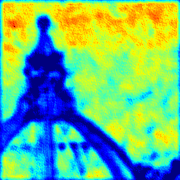}\\
\end{minipage}
\begin{minipage}{0.22\linewidth}
\centering
\includegraphics[width=1\linewidth]{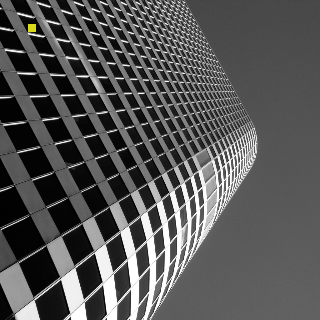}\\
\includegraphics[width=1\linewidth]{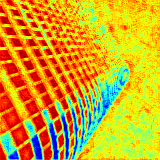}\\
\includegraphics[width=1\linewidth]{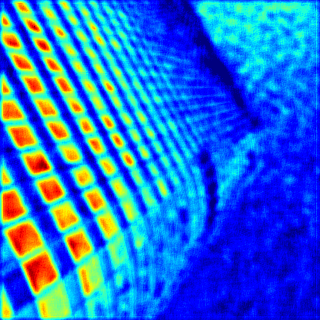}\\
\end{minipage}
\begin{minipage}{0.22\linewidth}
\centering
\includegraphics[width=1\linewidth]{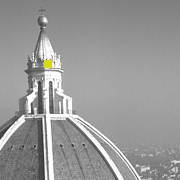}\\
\includegraphics[width=1\linewidth]{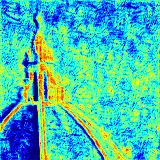}\\
\includegraphics[width=1\linewidth]{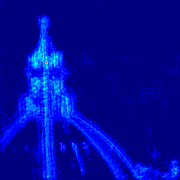}\\
\end{minipage}
\begin{minipage}{0.22\linewidth}
\centering
\includegraphics[width=1\linewidth]{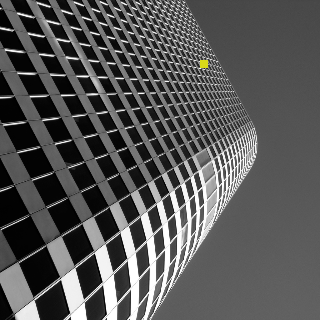}\\
\includegraphics[width=1\linewidth]{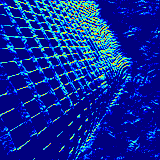}\\
\includegraphics[width=1\linewidth]{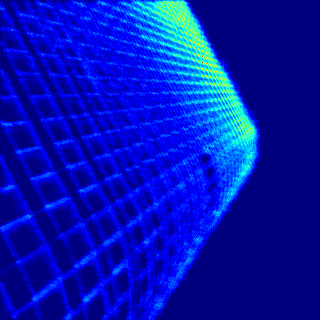}\\
\end{minipage}
\centering
\caption{Visualization of our patch-wise non-local attention operation and its counterpart (non-local neural networks \cite{nonlocalnet}) which is represented as NLNet. The original images are shown in the first row and the query pixel/patch is labeled with a yellow dotted box. Long-range dependence (LRD) is shown below in the form of heat-maps. The brighter color indicates higher engagement. }
\label{im_nlatt} 
\end{figure}

\textbf{Patch-wise Non-local Attention Model.} Based on the non-local neural networks \cite{nonlocalnet} and existing patch-wise non-local operations, we propose a novel patch-wise non-local attention model. As illustrated in Fig.~\ref{im_nonlocal}, given the input feature maps $\mathbf{F}^{i}_{FEM} \in \mathbb{R}^{C\times W \times H}$, we use three independent $1\times 1$ convolutional layers with trainable parameters $\mathbf{W}_{g }$, $\mathbf{W}_{\theta}$, and $\mathbf{W}_{\varphi}$ as the embedding functions. These three embedding functions are represented as $g$, $\theta$, and $\varphi$. They are used to generate 
the query ($\mathbf{Q}$), key ($\mathbf{K}$) and value ($\mathbf{V}$). The embedding process does not change the size of feature maps, which can be defined as follows:
\begin{equation}
    \begin{cases}
 & \mathbf{Q} = \theta(\mathbf{F}^{i}_{FEM})\\ 
 & \mathbf{K} = \varphi(\mathbf{F}^{i}_{FEM})\\
 & \mathbf{V} = g(\mathbf{F}^{i}_{FEM}). 
\end{cases}
\end{equation}

Rather than directly reshaping $\mathbf{Q}, \mathbf{K}$ and calculating the relationship in pixel level as \cite{nonlocalnet}, we propose to utilize the \textbf{unfold} operation to extract sliding local patches from these transformed feature maps with stride $s$ and patch-size of $W_p\times H_p$. Then, as shown in Fig.~\ref{im_nonlocal}, we obtain three groups of 3D patches. Each group has $N$ patches with the size of each patch being $C\times W_p\times H_p$. The novelty of our proposed non-local attention model is to calculate the relationship in the unit of 3D patch, which is more effective and robust.

Next, we reshape each 3D patch unfolded from $\mathbf{Q}$ and $\mathbf{K}$ into a 1D feature vector, obtaining $\tilde{\mathbf{Q}}$ and $\tilde{\mathbf{K}}$. The distance matrix denoted by $\mathbf{M}\in \mathbb{R}^{N\times N}$ can be efficiently calculated by dot product: 
\begin{equation}
\phi(\tilde{\mathbf{Q}},\tilde{\mathbf{K}}) = softmax(\tilde{\mathbf{Q}}^{T}\tilde{\mathbf{K}}).
\end{equation}

Each 3D patch in $\mathbf{V}$ can aggregate useful information to update itself guided by this distance matrix $\mathbf{M}$, which is essentially a weighted sum updating process in the unit of patch.

Finally, we utilize the \textbf{fold} operation to combine this array of updated sliding local patches into a feature map with the size of $C\times W\times H$. This process can be viewed as the inverse process of the unfold operation. In consideration of overlapping between patches, we apply the averaging to deal with the overlapped areas. Compared with \cite{irpnet} applying a prediction neural network to infer the similarity between pairwise patches and \cite{n3net} using a differentiable \textit{k}-Nearest method to perform patch matching in a limited search region, our proposed patch-wise non-local attention model is more effective to fully utilize useful information from the whole image to restore each patch.

The pixel-wise non-local method \cite{nonlocalnet} (corresponding to Fig.~\ref{im_nl_cp} (a)) is the counterpart of our method, which has been widely used in many image restoration tasks \cite{rnan,n3net}. To demonstrate that our proposed patch-wise non-local method can capture more reliable long-range dependence than \cite{nonlocalnet} in the same case, we visualize their non-local attention maps in image denoising task ($\bm{\sigma}=30$), and the comparison is presented in Fig.~\ref{im_nlatt}. The heat-maps of our method are extracted from the second CAB of COLA-Net, and the results of its counterpart are extracted from the same place by replacing our patch-wise non-local method with the non-local neural networks \cite{nonlocalnet}. Specifically, in image degradation, reliable correlations can aggregate more useful information to improve restoration results. From Fig.~\ref{im_nlatt}, it is obvious that the long-range dependence built based on pixels is susceptible to noisy signals. In comparison, our learnable patch-wise non-local attention model is scarcely influenced by noisy signals and can capture more reliable long-range dependence.

\begin{table}[H]
    \centering
        \caption{Computational Complexity and Parameter Number Comparison. PSNR and SSIM values are obtained from Urban100 test set \cite{urban100} with noise level equaling to 50.}
    \begin{tabular}{|c|c|c|c|}

      \hline
         Algorithm & \#Parameter $\downarrow$ &Running Time $\downarrow$ &PSNR/SSIM $\uparrow$ \\
         \hline
           \hline
         DnCNN \cite{dncnn}&0.56M& \textbf{0.03}s&26.28/0.7874\\
         ADNet \cite{dilated2}& \textbf{0.52}M&0.06s&26.64/0.8072\\
         N3Net \cite{n3net}&0.72M&0.08s&26.82/0.8184\\
         NLRN \cite{nlrn}&0.35M&38.17s&27.49/0.8279\\
         RNAN \cite{rnan}&8.96M&1.25s&27.65/0.8348\\
         \hline \hline
         COLA-B& 1.10M&0.51s&27.76/0.8373\\
           \hline
         COLA-E& 1.88M&0.68s&\textbf{27.84}/\textbf{0.8392}\\
           \hline
    \end{tabular}
    \label{tb_cplx}
\end{table}

\begin{table*}[t]
\caption{Quantitative results (PSNR and SSIM) about gray-scale image denoising. Best results are \textbf{highlighted}.}
\centering
\begin{tabular}{|c|c|c|c|c|c|c|c|c|c|}
\hline
Dataset & $\bm{\sigma}$ & BM3D \cite{BM3D} & DnCNN \cite{dncnn} & FFDNet \cite{ffdnet} & ADNet \cite{dilated2} & N3Net \cite{n3net} & NLRN \cite{nlrn} & COLA-B & COLA-E\\
\hline \hline
\multirow{4}*{Set12} & 15 & 32.37/0.8952 & 32.86/0.9031 & 32.75/0.9027 & 32.98/0.9044 & 33.03/0.9056 & 33.16/0.9070 & \underline{33.20}/\underline{0.9088}& \textbf{33.27}/\textbf{0.9097} \\
& 25 & 29.96/0.8504 & 30.44/0.8622 & 30.43/0.8634 & 30.58/0.8649 & 30.55/0.8648 & 30.80/0.8689 & \underline{30.85}/\underline{0.8701}& \textbf{30.90}/\textbf{0.8716}\\
& 50 & 26.70/0.7676 & 27.19/0.7829 & 27.31/0.7903 & 27.37/0.7903 & 27.43/0.7948 & 27.64/0.7980 & \underline{27.73}/\underline{0.8020}& \textbf{27.77}/\textbf{0.8032}\\
& 70 & 25.21/0.7176 & 25.56/0.7273 & 25.81/0.7451 & 25.63/0.7364 & 25.90/0.7510 & -/- & \underline{26.18}/\underline{0.7572}& \textbf{26.25}/\textbf{0.7606}\\
\hline
\hline
\multirow{4}*{BSD68} & 15 & 31.07/0.8717 & 31.73/0.8907 & 31.63/0.8902 & 31.74/0.8910 & 31.78/0.8927 & 31.88/0.8932 & \underline{31.88}/\underline{0.8938} & \textbf{31.92}/\textbf{0.8968}\\
& 25 & 28.57/0.8013 & 29.23/0.8278 & 29.19/0.8289 & 29.25/0.8288 & 29.30/0.8321 & 29.41/0.8331 & \underline{29.42}/\underline{0.8339}& \textbf{29.46}/\textbf{0.8368}\\
& 50 & 25.62/0.6864 & 26.23/0.7189 & 26.29/0.7345 & 26.29/0.7210 & 26.39/0.7293 & 26.47/0.7298 & \underline{26.48}/\underline{0.7308}& \textbf{26.52}/\textbf{0.7340}\\
& 70 & 24.46/0.6323 & 24.85/0.6567 & 25.04/0.6700 & 24.95/0.6625 & 25.14/\underline{0.6753} & -/- & \underline{25.17}/0.6735& \textbf{25.25}/\textbf{0.6788}\\
\hline
\hline
\multirow{4}*{Urban100} & 15 & 32.35/0.9220 & 32.68/0.9255 & 32.43/0.9273 & 32.87/0.9304 & 33.08/0.9333 & 33.45/0.9354 & \underline{33.60}/\underline{0.9376} & \textbf{33.73}/\textbf{0.9387}\\
& 25 & 29.71/0.8777 & 29.97/0.8797 & 29.92/0.8887 & 30.24/0.8920 & 30.19/0.8925 & 30.94/0.9018 & \underline{31.17}/\underline{0.9062}& \textbf{31.33}/\textbf{0.9086}\\
& 50 & 25.95/0.7791 & 26.28/0.7874 & 26.52/0.8057 & 26.64/0.8072 & 26.82/0.8184 & 27.49/0.8279 & \underline{27.76}/\underline{0.8373}& \textbf{27.84}/\textbf{0.8392}\\
& 70 & 24.27/0.7165 & 24.34/0.7178 & 24.87/0.7495 & 24.53/0.7304 & 25.15/0.7658 & -/- & \underline{25.99}/\underline{0.7848}& \textbf{26.15}/\textbf{0.7910}\\
\hline
\hline
\multicolumn{2}{|c|}{Parameters}& - & 0.56M & 0.49M & 0.52M & 0.72M & 0.35M & 1.10M&1.88M\\
\hline
\end{tabular}
\label{tb_sdn}
\end{table*}
\section{Experimental Results}
\subsection{Training Details and Evaluation}
To verify the effectiveness of our proposed COLA-Net, we apply the basic version (COLA-B) and the enhanced version (COLA-E) of COLA-Net into three typical image restoration tasks: synthetic image denoising, real image denoising, and compression artifact reduction. For synthetic image denoising and compression artifact reduction tasks, we train COLA-Net with DIV2K \cite{div2k} dataset, which contains 800 high-quality images. For real image denoising, we adopt training data containing 400 images corrupted by synthetic noise from BSD500 \cite{bsd500} dataset with the noise-level randomly selected from [10,50] and 120 images corrupted by realistic noise from RENOIR \cite{renoir} dataset, which is similar to CBDNet \cite{cbdnet}. Our model is trained on a GTX1080Ti GPU with the initial learning rate $lr = 1\times10^{-3}$ and performs halving per 200 epochs. During training, we employ Adam optimizer and each mini-batch contains 32 images with size of $64\times64$ randomly cropped from training data. A data augmentation method, which is the same as RNAN \cite{rnan} is also applied in the training process. For each task, we apply commonly used test sets for testing and report PSNR (dB) and/or SSIM \cite{ssim} to evaluate the performance of each method.

\begin{table}[h]
\caption{PSNR and SSIM comparison with RNAN \cite{rnan} on gray-scale image denoising. Best results are \textbf{highlighted}.}
\begin{center}
\begin{tabular}{|c|c|c|c|c|}
\hline
Dataset                    & $\bm{\sigma}$  & RNAN \cite{rnan}        & COLA-B         & COLA-E \\ \hline \hline
\multirow{4}{*}{BSD68}     & 10  & \underline{34.04}/\underline{0.9295} & 34.03/0.9290 & \textbf{34.08}/\textbf{0.9299}    \\
                           & 30  & 28.61/0.8094 & \underline{28.62}/\underline{0.8096} & \textbf{28.64}/\textbf{0.8110}    \\
                           & 50  & 26.48/0.7306 & \underline{26.48}/\underline{0.7308} & \textbf{26.52}/\textbf{0.7340}    \\
                           & 70  & \underline{25.18}/\underline{0.6746} & 25.17/0.6735 & \textbf{25.25}/\textbf{0.6788}    \\ \hline \hline
\multirow{4}{*}{Urban100}  & 10  & 35.52/0.9553 & \underline{35.56}/\underline{0.9554} & \textbf{35.64}/\textbf{0.9559}    \\
                           & 30  & 30.20/0.8902 & \underline{30.26}/\underline{0.8910} & \textbf{30.41}/\textbf{0.8936}    \\
                           & 50  & 27.65/0.8348 & \underline{27.76}/\underline{0.8373} & \textbf{27.84}/\textbf{0.8392}    \\
                           & 70  & 25.89/0.7835 & \underline{25.99}/\underline{0.7848} & \textbf{26.15}/\textbf{0.7910}    \\ \hline \hline
\multicolumn{2}{|c|}{Parameters} & 8.96M         & 1.10M         & 1.88M      \\ \hline
\end{tabular}
\end{center}
\label{tb_cprnan}
\end{table}

\begin{figure*}[htbp]
\centering
\small 
\begin{minipage}{0.17\linewidth}
\centering
\includegraphics[width=.98\columnwidth,height=3.55cm]{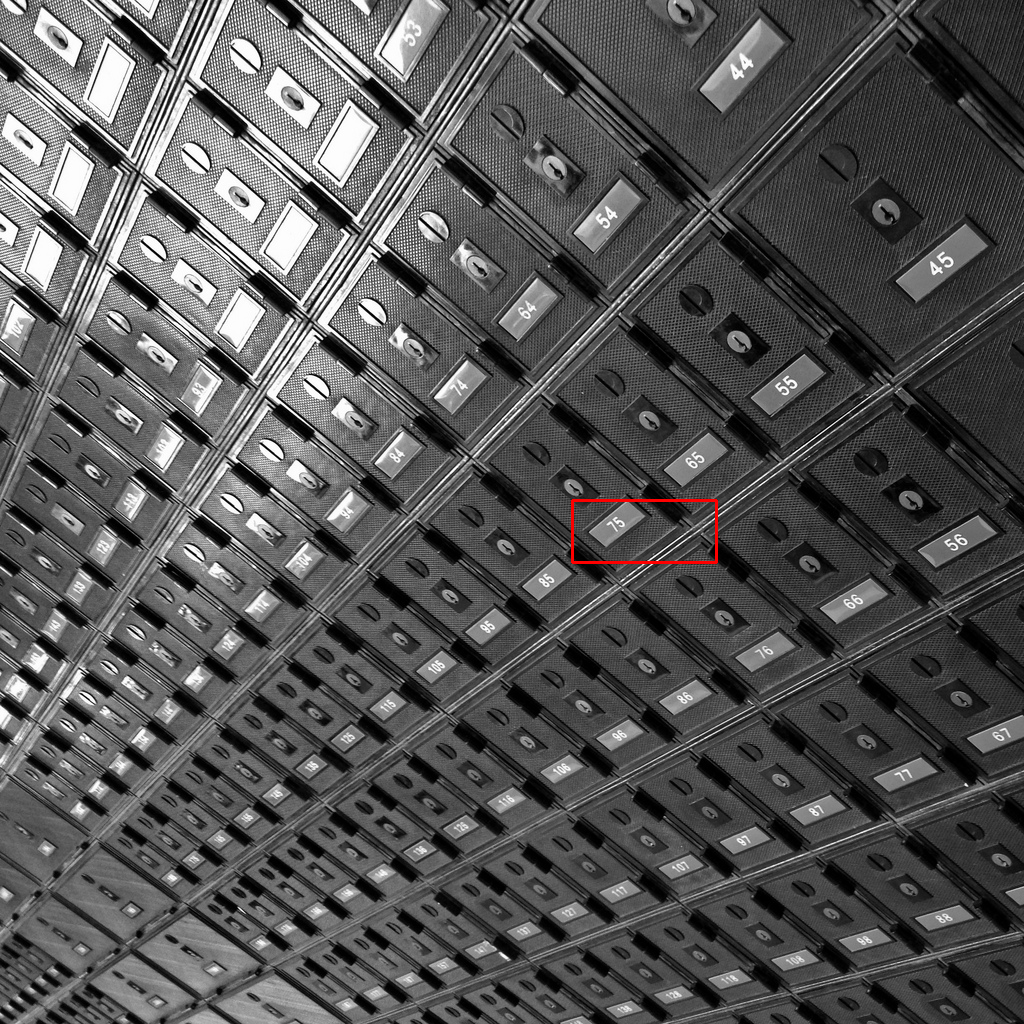}\\Urban100: img006\\~\\
\end{minipage}
\begin{minipage}{0.16\linewidth}
\centering
\includegraphics[width=.98\columnwidth,height=1.4cm]{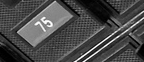}\\HQ\\(PSNR/SSIM)\\
\includegraphics[width=.95\columnwidth,height=1.4cm]{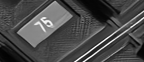}\\ADNet\\(27.44/0.8449)
\end{minipage}
\begin{minipage}{0.16\linewidth}
\centering
\includegraphics[width=.98\columnwidth,height=1.4cm]{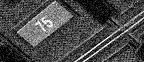}\\Noisy\\(20.60/0.5748)\\
\includegraphics[width=.98\columnwidth,height=1.4cm]{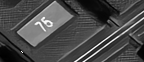}\\N3Net\\(27.40/0.8457)
\end{minipage}
\begin{minipage}{0.16\linewidth}
\centering
\includegraphics[width=.98\columnwidth,height=1.4cm]{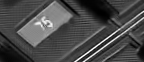}\\BM3D\\(25.88/0.7969)\\
\includegraphics[width=.95\columnwidth,height=1.4cm]{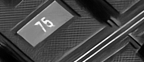}\\NLRN\\(28.46/0.8688)
\end{minipage}
\begin{minipage}{0.16\linewidth}
\centering
\includegraphics[width=.98\columnwidth,height=1.4cm]{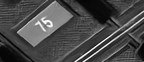}\\DnCNN\\(26.38/0.8052)\\
\includegraphics[width=.98\columnwidth,height=1.4cm]{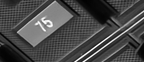}\\COLA-B (Ours)\\(29.16/0.8951)
\end{minipage}
\begin{minipage}{0.16\linewidth}
\centering
\includegraphics[width=.98\columnwidth,height=1.4cm]{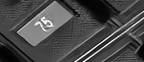}\\FFDNet\\(27.05/0.8359)\\
\includegraphics[width=.98\columnwidth,height=1.4cm]{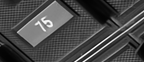}\\COLA-E (Ours)\\(\textbf{29.29/0.8964})
\end{minipage}\\~\\~\\
\begin{minipage}{0.17\linewidth}
\centering
\includegraphics[width=.98\columnwidth,height=3.55cm]{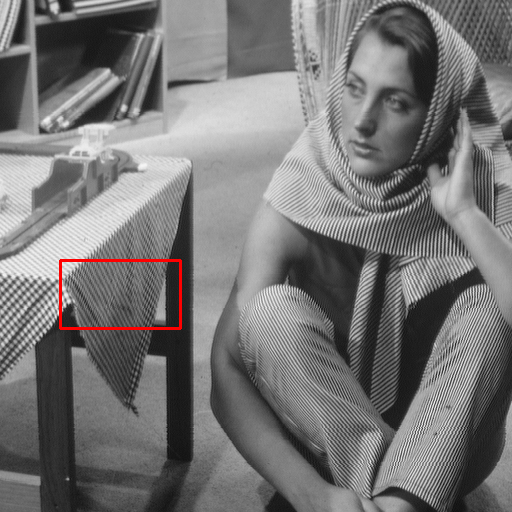}\\Set12: Barbara\\~\\
\end{minipage}
\begin{minipage}{0.16\linewidth}
\centering
\includegraphics[width=.98\columnwidth,height=1.4cm]{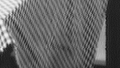}\\HQ\\(PSNR/SSIM)\\
\includegraphics[width=.98\columnwidth,height=1.4cm]{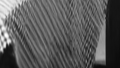}\\ADNet\\(30.22/0.8840)
\end{minipage}
\begin{minipage}{0.16\linewidth}
\centering
\includegraphics[width=.98\columnwidth,height=1.4cm]{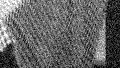}\\Noisy\\(20.28/0.4040)\\
\includegraphics[width=.98\columnwidth,height=1.4cm]{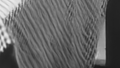}\\N3Net\\(30.20/0.8850)
\end{minipage}
\begin{minipage}{0.16\linewidth}
\centering
\includegraphics[width=.98\columnwidth,height=1.4cm]{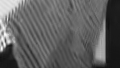}\\BM3D\\(27.08/0.8070)\\
\includegraphics[width=.98\columnwidth,height=1.4cm]{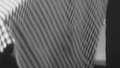}\\NLRN\\(30.99/0.8974)
\end{minipage}
\begin{minipage}{0.16\linewidth}
\centering
\includegraphics[width=.98\columnwidth,height=1.4cm]{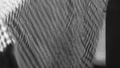}\\DnCNN\\(29.21/0.8619)\\
\includegraphics[width=.98\columnwidth,height=1.4cm]{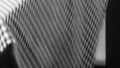}\\COLA-B (Ours)\\(31.45/0.9051)
\end{minipage}
\begin{minipage}{0.16\linewidth}
\centering
\includegraphics[width=.98\columnwidth,height=1.4cm]{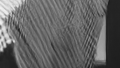}\\FFDNet\\(29.98/0.8790)\\
\includegraphics[width=.98\columnwidth,height=1.4cm]{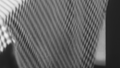}\\COLA-E (Ours)\\(\textbf{31.50/0.9058})
\end{minipage}
\caption{Visual comparison for gray image denoising of various methods on two samples from Urban100 and Set12 with noise level $\sigma =25$.}
\label{im_urban}
\end{figure*}

\begin{table*}[t]
\caption{Quantitative results (PSNR and SSIM) of compression artifact reduction. Best results are \textbf{highlighted}.}
\begin{tabular}{|c|c|c|c|c|c|c|c|c|c|}
\hline 
Dataset &$q$ & JPEG & SA-DCT \cite{sadct} & ARCNN \cite{arcnn} & TNRD \cite{tnrd} & DnCNN \cite{dncnn} & RNAN \cite{rnan} & COLA-B& COLA-E\\
\hline \hline
\multirow{4}*{LIVE1} & 10 & 27.77/0.7905 & 28.86/0.8093 & 28.98/0.8076 & 29.15/0.8111 & 29.19 /0.8123& \underline{29.63}/\underline{0.8239} & 29.60/0.8226&\textbf{29.66}/\textbf{0.8234}\\
& 20 & 30.07/0.8683 & 30.81/0.8781 & 31.29/0.8733 & 31.46/0.8769 & 31.59/0.8802 & \underline{32.03}/\underline{0.8877} & 31.98/0.8865&\textbf{32.06}/\textbf{0.8880}\\
& 30 & 31.41/0.9000 & 32.08/0.9078 & 32.69/0.9043 & 32.84/0.9059 & 32.98/0.9090 & \underline{33.45}/\underline{0.9149} & 33.40/0.9141 &\textbf{33.48}/\textbf{0.9152}\\
& 40 & 32.35/0.9173 & 32.99/0.9240 & 33.63/0.9198 & -/- & 33.96/0.9247 & \underline{34.47}/\underline{0.9299} & 34.43/0.9293&\textbf{34.49}/\textbf{0.9300}\\
\hline
\hline
\multirow{4}*{Classic5} & 10 & 27.82/0.7800 & 28.88/0.8071 & 29.04/0.7929 & 29.28/0.7992 & 29.40/0.8026 & 29.96/0.8178 & \underline{29.96}/\underline{0.8180}&\textbf{30.03}/\textbf{0.8184}\\
& 20 & 30.12/0.8541 & 30.92/0.8663 & 31.16/0.8517 & 31.47/0.8576 & 31.63/0.8610 & 32.11/0.8693 & \underline{32.18}/\underline{0.8695} &\textbf{32.28}/\textbf{0.8706}\\
& 30 & 31.48/0.8844 & 32.14/0.8914 & 32.52/0.8806 & 32.74/0.8837 & 32.91/0.8861 & 33.38/0.8924 & \underline{33.48}/\underline{0.8929}&\textbf{33.54}/\textbf{0.8935}\\
& 40 & 32.43/0.9011 & 33.00/0.9055 & 33.34/0.8953 & -/- & 33.77/0.9003 & 34.27/0.9061 & \underline{34.33}/\underline{0.9063}&\textbf{34.38}/\textbf{0.9066}\\
\hline
\hline
\multicolumn{2}{|c|}{Parameters}&-&-&0.12M&-&0.56M&8.96M&1.10M&1.88M\\
\hline
\end{tabular}
\label{tb_db}
\end{table*}

\begin{figure*}[h]
\centering
\small 
\begin{minipage}[t]{0.119\linewidth}
\centering
\includegraphics[width=1\columnwidth,height=2.2cm]{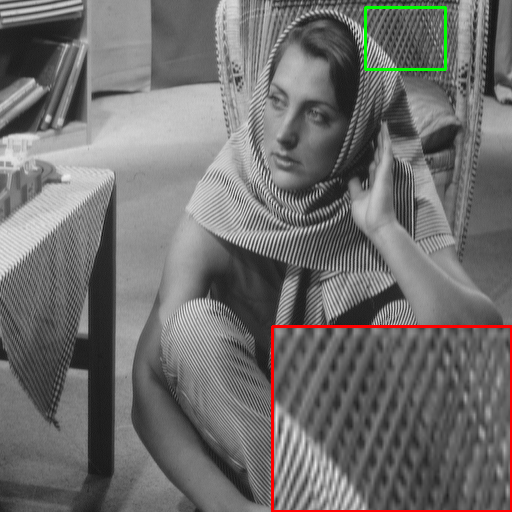}\\HQ (Classic5: Barbara)\\~\\
\includegraphics[width=1\columnwidth,height=2.2cm]{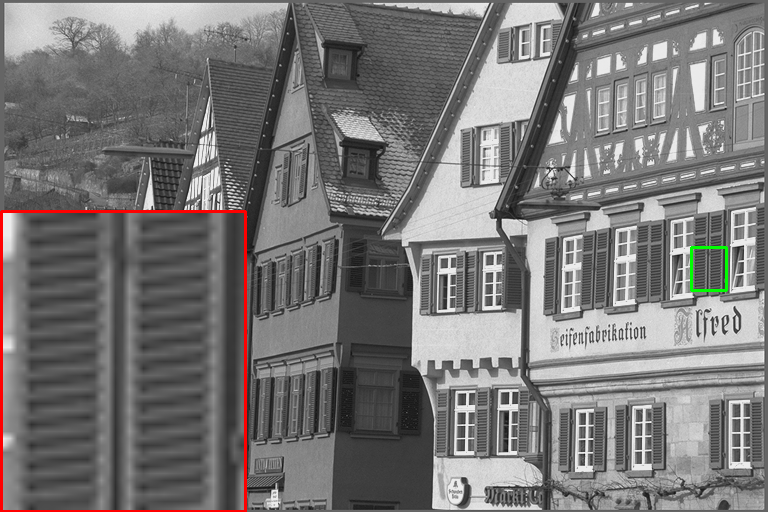}\\HQ (LIVE1: Buildings)
\end{minipage}%
\hspace{0.01mm}
\begin{minipage}[t]{0.12\linewidth}
\centering
\includegraphics[width=.98\columnwidth,height=2.2cm]{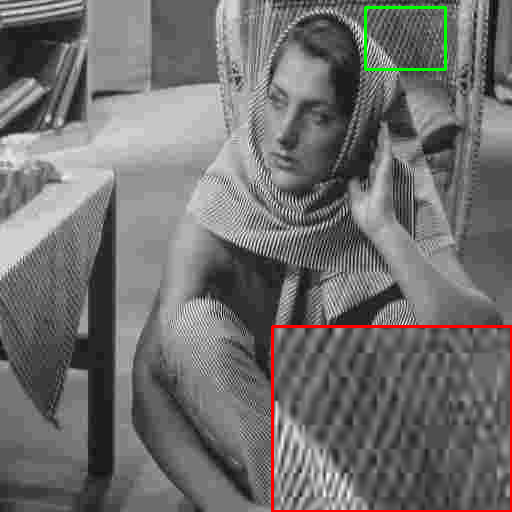}\\JPEG (q=10)\\25.78/0.7621\\~\\
\includegraphics[width=.98\columnwidth,height=2.2cm]{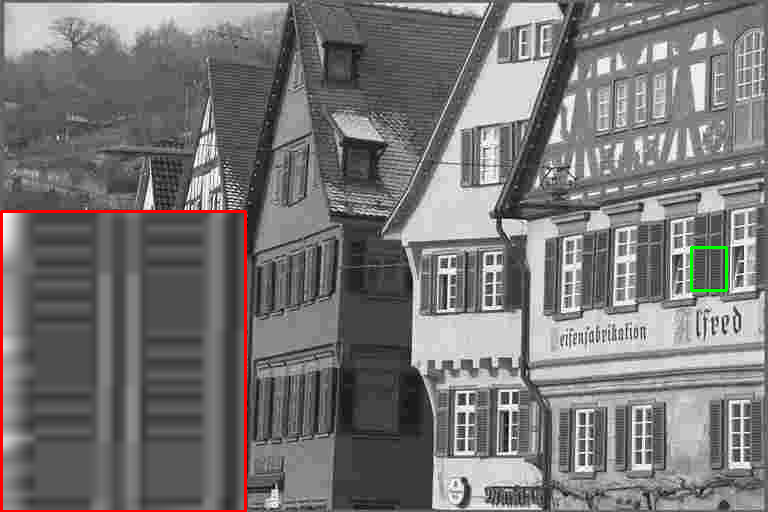}\\JPEG (q=10)\\25.07/0.7632
\end{minipage}%
\hspace{0.01mm}
\begin{minipage}[t]{0.12\linewidth}
\centering
\includegraphics[width=1\columnwidth,height=2.2cm]{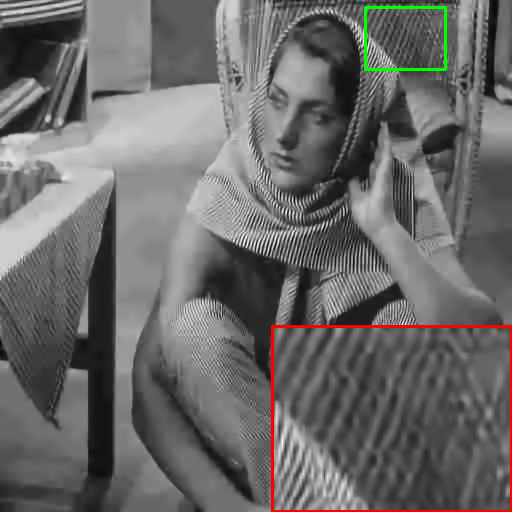}\\ARCNN \cite{arcnn}\\26.84/0.7980\\~\\
\includegraphics[width=1\columnwidth,height=2.2cm]{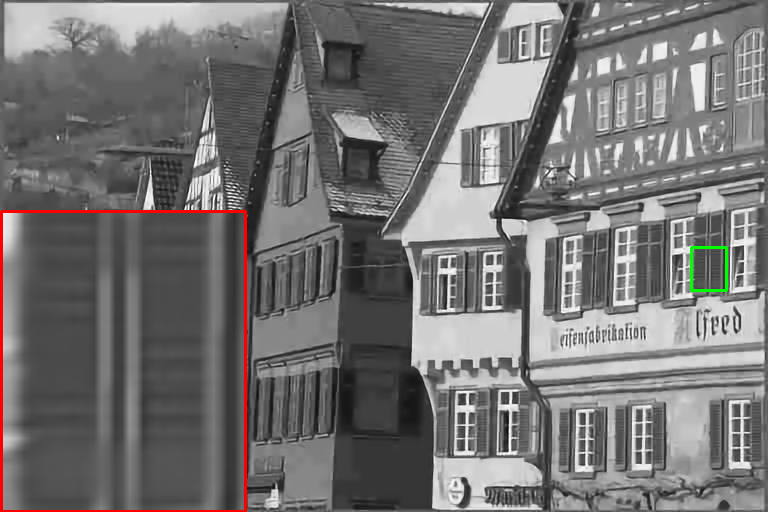}\\ARCNN \cite{arcnn}\\26.32/0.7976
\end{minipage}
\begin{minipage}[t]{0.12\linewidth}
\centering
\includegraphics[width=1\columnwidth,height=2.2cm]{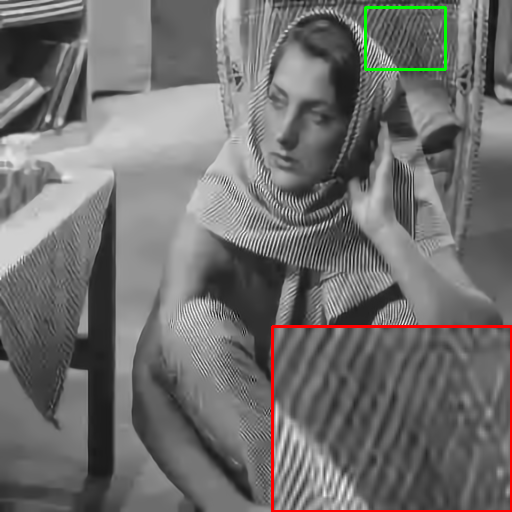}\\TNRD \cite{tnrd}\\27.23/0.8099\\~\\
\includegraphics[width=1\columnwidth,height=2.2cm]{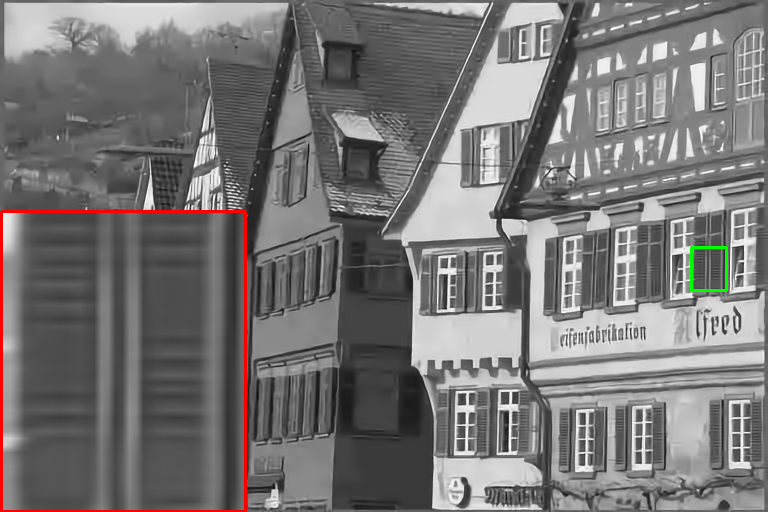}\\TNRD \cite{tnrd}\\26.64/0.8055
\end{minipage}
\begin{minipage}[t]{0.12\linewidth}
\centering
\includegraphics[width=1\columnwidth,height=2.2cm]{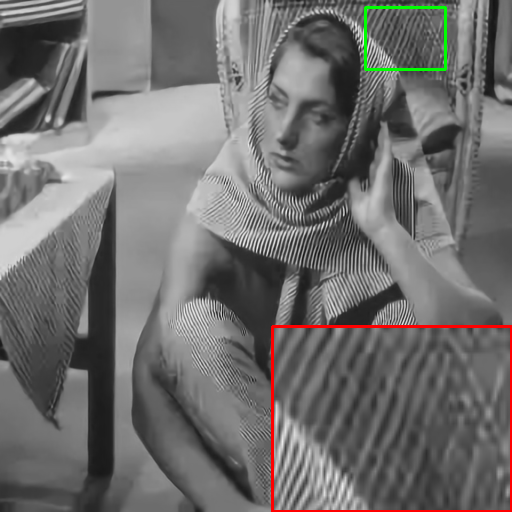}\\DnCNN \cite{dncnn}\\27.59/0.8161\\~\\
\includegraphics[width=1\columnwidth,height=2.2cm]{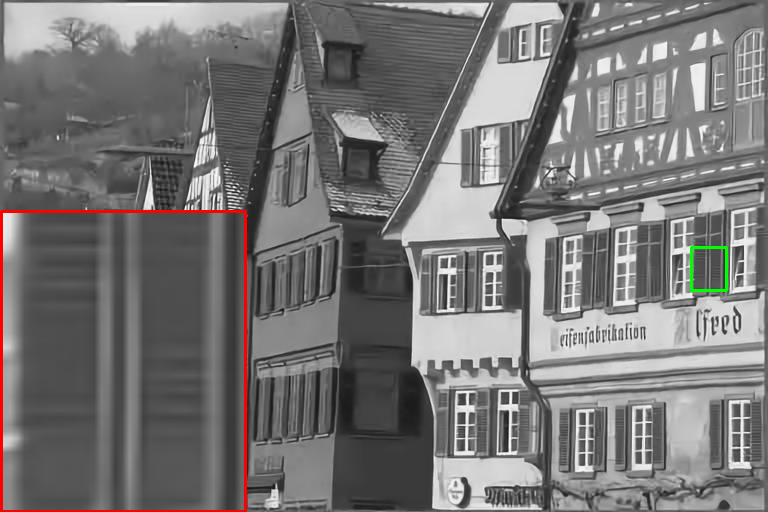}\\DnCNN \cite{dncnn}\\26.75/0.8066
\end{minipage}
\begin{minipage}[t]{0.12\linewidth}
\centering
\includegraphics[width=1\columnwidth,height=2.2cm]{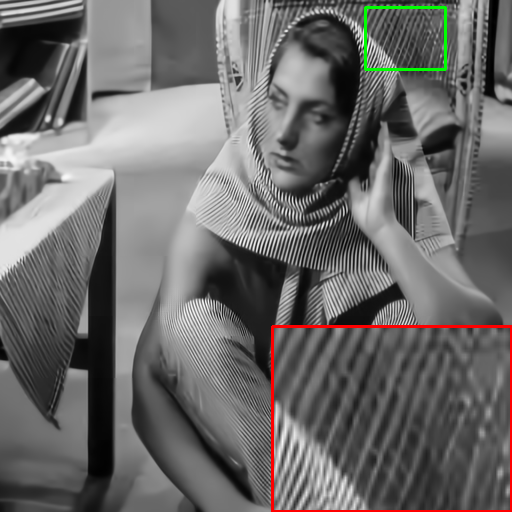}\\RNAN \cite{rnan}\\28.86/0.8516\\~\\
\includegraphics[width=1\columnwidth,height=2.2cm]{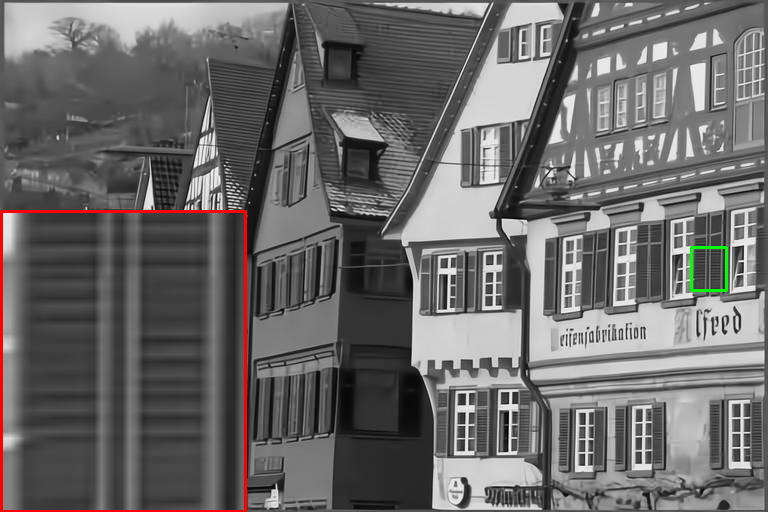}\\RNAN \cite{rnan}\\27.58/0.8314
\end{minipage}
\begin{minipage}[t]{0.12\linewidth}
\centering
\includegraphics[width=1\columnwidth,height=2.2cm]{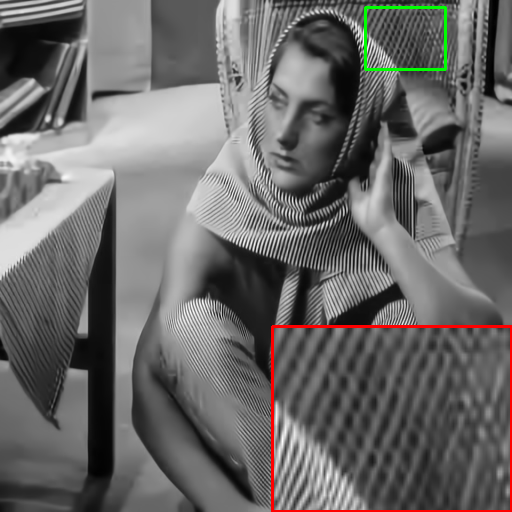}\\COLA-B (Ours)\\29.04/0.8549\\~\\
\includegraphics[width=1\columnwidth,height=2.2cm]{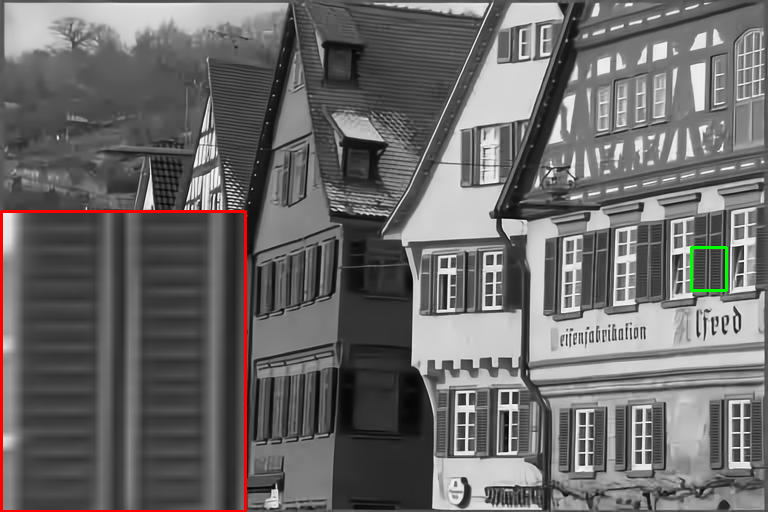}\\COLA-B (Ours)\\27.63/0.8330
\end{minipage}
\begin{minipage}[t]{0.12\linewidth}
\centering
\includegraphics[width=1\columnwidth,height=2.2cm]{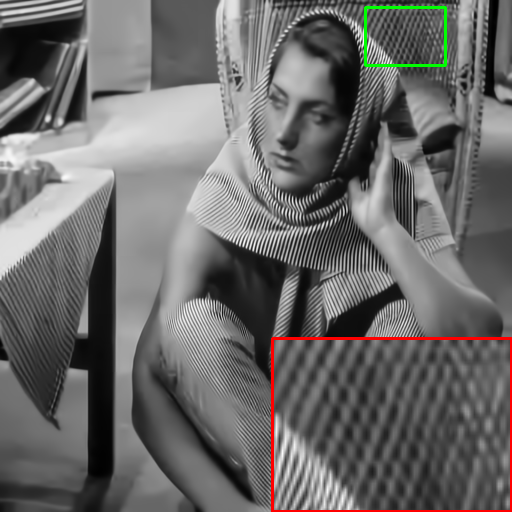}\\COLA-E (Ours)\\\textbf{29.20/0.8581}\\~\\
\includegraphics[width=1\columnwidth,height=2.2cm]{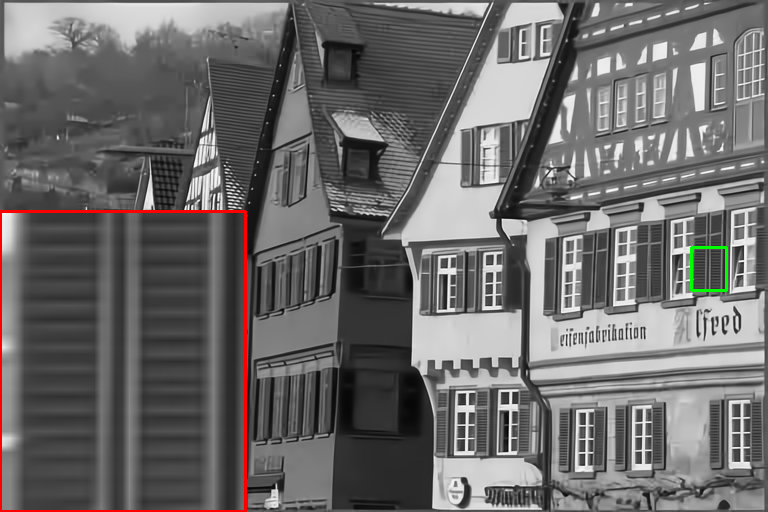}\\COLA-E (Ours)\\\textbf{27.75/0.8352}
\end{minipage}
\vspace{-5pt}
\centering
\caption{Visual comparison of image compression artifact reduction application of various methods with JPEG quality q = 10.}
\label{im_car} 
\end{figure*}

\subsection{Comparisons with State-of-the-Art Methods}
We first make a brief complexity analysis of some representative methods and then compare our proposed COLA-Net with some recent state-of-the-art methods in synthetic image denoising, real image denoising, and compression artifact reduction applications.

\subsubsection{\textbf{Computational Complexity}}
Realizing that the number of trainable parameters can not completely reflect the complexity of a model, especially for deep non-local methods. Thus, we employ both the number of trainable parameters and the running time that a model takes to process a $256\times256$ image to evaluate the complexity of different methods. Note that all running times are tested on GPU. We also provide the denoising performance ($\sigma=50$) on the Urban100 \cite{urban100} dataset of various methods. The complexity analysis of several representative methods is reported in Table~\ref{tb_cplx}. Compared with some competitive non-local-based denoiser, \textit{e.g.}, RNAN \cite{rnan}, which stacked a lot of convolutional layers with the help of residual connection, and NLRN \cite{nlrn}, which applied a recurrent strategy to cycle through one input many times, both the basic version (COLA-B) and enhanced version (COLA-E) of our COLA-Net have an attractive complexity and achieve quite promising performance.

\subsubsection{\textbf{Synthetic Image Denoising}}
For this application, three standard benchmark datasets, i.e., Set12, BSD68 \cite{bsd68}, and Urban100 \cite{urban100} are evaluated. We compare the two versions of our COLA-Net with some state-of-the-art denoising methods, including some well-known denoisers, \textit{e.g.}, DnCNN \cite{dncnn} and FFDNet \cite{ffdnet}, and recent local attention-based method ADNet \cite{dilated2} as well as some competitive non-local denoisers such as BM3D \cite{BM3D}, N3Net \cite{n3net} and NLRN\cite{nlrn}. Additive white Gaussian noise (AWGN) with different noise levels (15, 25, 50, 70) is added to the clean images. The quantitative results (PSNR and SSIM ) are shown in Table~\ref{tb_sdn}, which clearly shows that the basic COLA-Net achieves the best performance in all three datasets, and the enhanced version can further improve the performance. It is worth noting that the superiority of our COLA-Net is more obvious in the case of strong noise. Especially for non-local methods, high-intensity noisy signals will disturb the construction of long-range dependence; thus, in this case, computing self-similarity based on pixels \cite{nlrn} or within a limited area \cite{n3net,irpnet} is not reliable. The visual comparisons of denoising results of different methods are shown in Fig.~\ref{im_urban}. One can observe that our COLA-Net produces higher visual quality than other methods. Specifically, the non-local denoising methods \cite{BM3D,nlrn} produce the restoration results with over-smooth artifacts, and the local attention method \cite{dilated2} output oversharp results. In comparison, our method generates more accurate textures, demonstrating the effectiveness of our proposed mixed feature attention mechanism.


To further prove the superiority of COLA-Net, we also compare our method with recent very deep non-local neural networks RNAN \cite{rnan}, with the quantitative results shown in Table \ref{tb_cprnan}. We can see that compared with RNAN, the basic COLA-Net obtains better performance in most cases, and the enhanced COLA-Net gets better performance in all test sets and noise levels. Simultaneously, the number of trainable parameters in two versions of COLA-Net is much lower than RNAN.

\subsubsection{\textbf{Compression Artifact Reduction}}
For this application, we compare our COLA-Net with some competitive methods: SA-DCT \cite{sadct}, ARCNN \cite{arcnn}, TNRD \cite{tnrd}, DnCNN \cite{dncnn}, and RNAN \cite{rnan}. The compressed images are generated by Matlab standard JPEG encoder with quality factor q = 10, 20, 30, 40. We evaluate the performance on LIVE1 \cite{live1} and Classic5 \cite{sadct} test sets, and the quantitative results are presented in Table~\ref{tb_db}. One can see that our proposed COLA-E achieves the best performance in all test sets under the evaluation of both PSNR and SSIM \cite{ssim}, and the COLA-B presents an attractive performance maintaining fewer parameters. Visual comparison of compression artifacts reduction is shown in Fig.~\ref{im_car}. We can find that our proposed model has better visual quality than other methods. Compared with the very deep non-local method \cite{rnan}, our image restoration results have fewer over-smooth artifacts, \textit{e.g.,} in the region of shutters and wicker chair. The superiority of our method benefits from the fact that our proposed COLA-Net can balance the contribution of local and non-local attention operations according to the characteristic of specific image content.

\begin{figure*}[h]
\centering
\small 
\begin{minipage}[t]{0.12\linewidth}
\centering
\includegraphics[width=1\columnwidth]{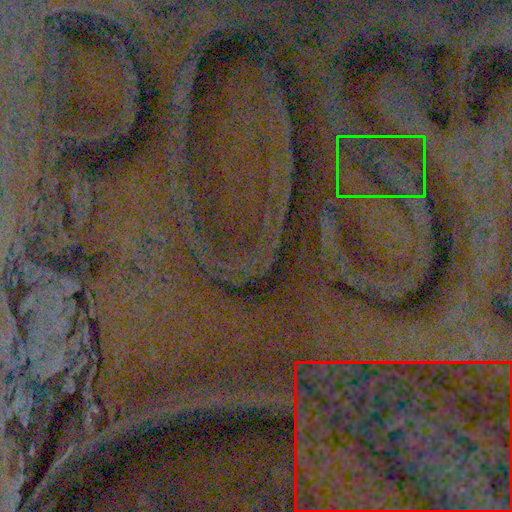}\\Noisy\\~\\
\includegraphics[width=1\columnwidth]{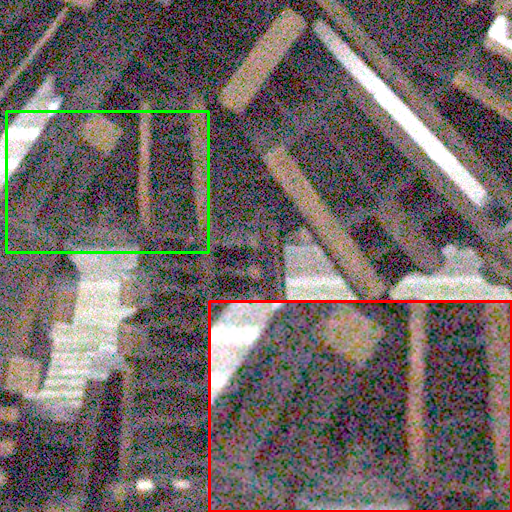}\\Noisy
\end{minipage}
\begin{minipage}[t]{0.12\linewidth}
\centering
\includegraphics[width=1\columnwidth]{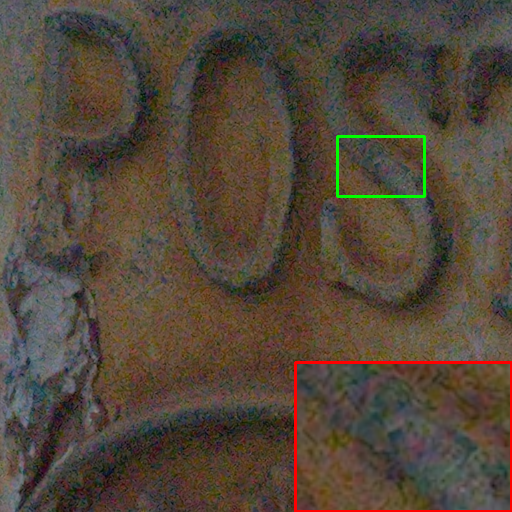}\\CDnCNN \cite{dncnn}\\~\\
\includegraphics[width=1\columnwidth]{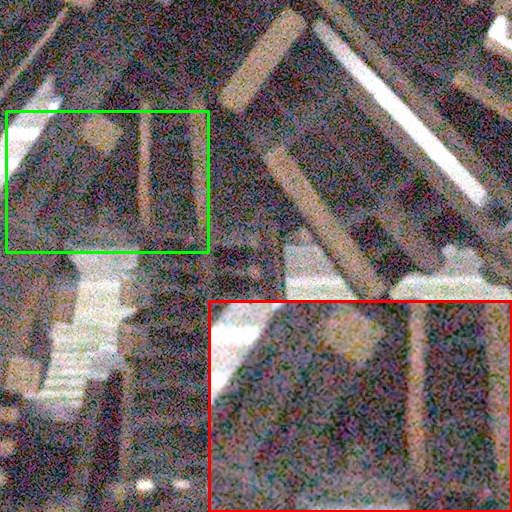}\\CDnCNN \cite{dncnn}
\end{minipage}
\begin{minipage}[t]{0.12\linewidth}
\centering
\includegraphics[width=1\columnwidth]{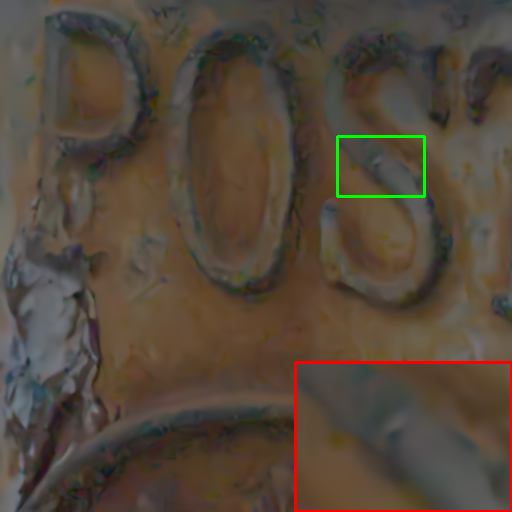}\\FFDNet \cite{ffdnet}\\~\\
\includegraphics[width=1\columnwidth]{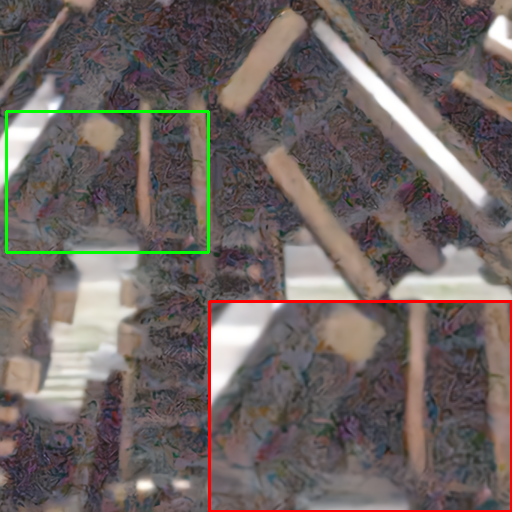}\\FFDNet \cite{ffdnet}
\end{minipage}
\begin{minipage}[t]{0.12\linewidth}
\centering
\includegraphics[width=1\columnwidth]{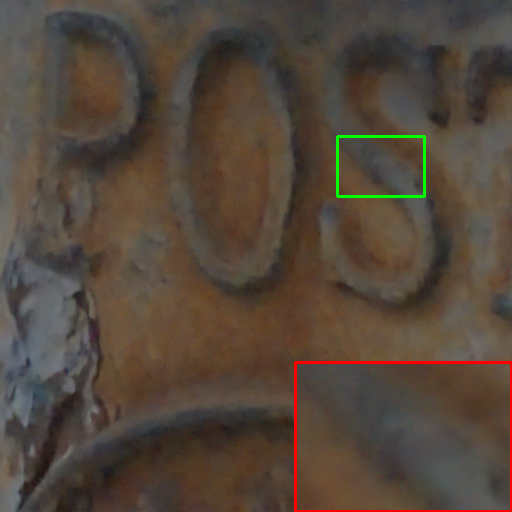}\\CBDNet \cite{cbdnet}\\~\\
\includegraphics[width=1\columnwidth]{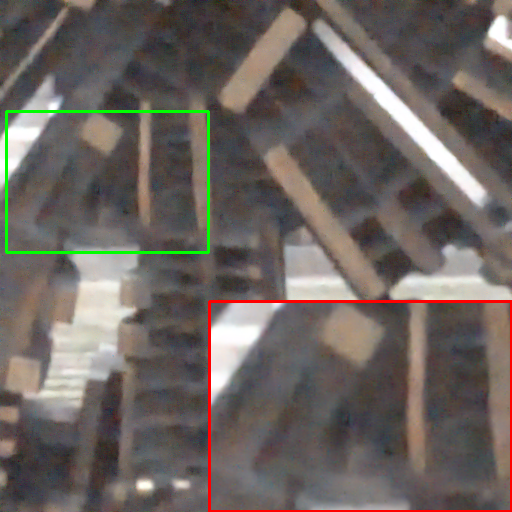}\\CBDNet \cite{cbdnet}
\end{minipage}
\begin{minipage}[t]{0.12\linewidth}
\centering
\includegraphics[width=1\columnwidth]{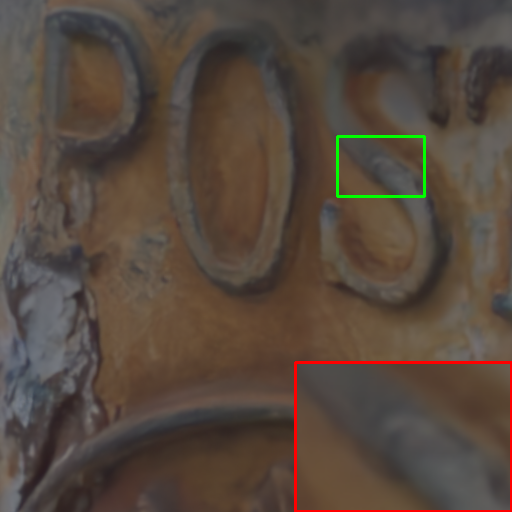}\\VDNet \cite{vdnet}\\~\\
\includegraphics[width=1\columnwidth]{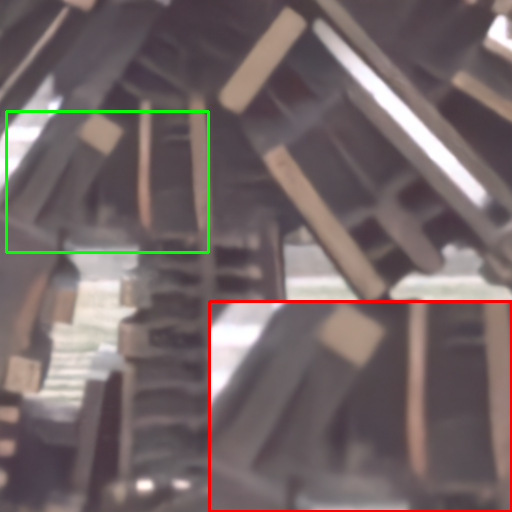}\\VDNet \cite{vdnet}
\end{minipage}
\begin{minipage}[t]{0.12\linewidth}
\centering
\includegraphics[width=1\columnwidth]{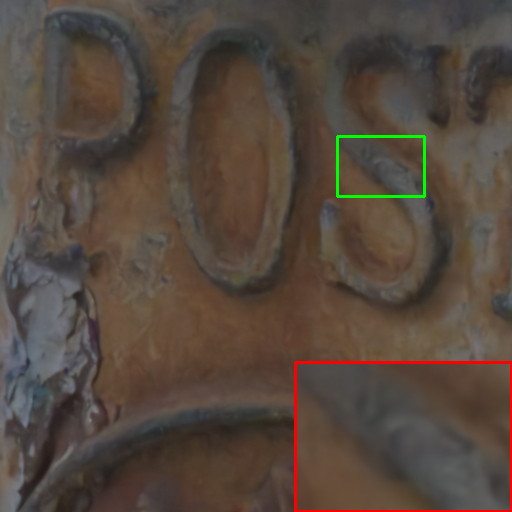}\\CycleISP \cite{cycleisp}\\~\\
\includegraphics[width=1\columnwidth]{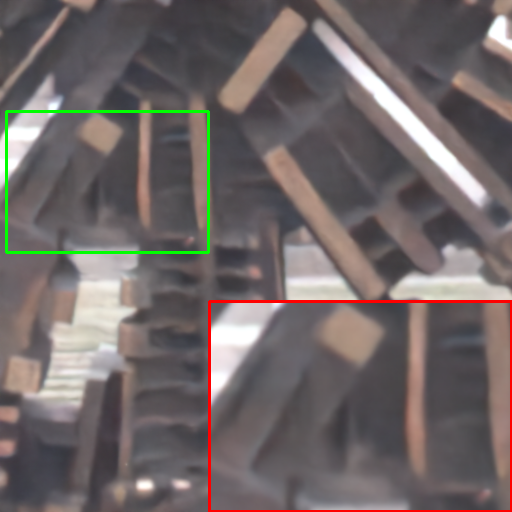}\\CycleISP \cite{cycleisp}
\end{minipage}
\begin{minipage}[t]{0.12\linewidth}
\centering
\includegraphics[width=1\columnwidth]{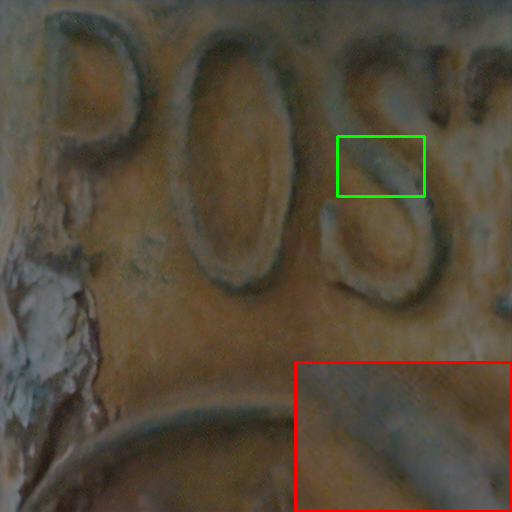}\\COLA-B (Ours)\\~\\
\includegraphics[width=1\columnwidth]{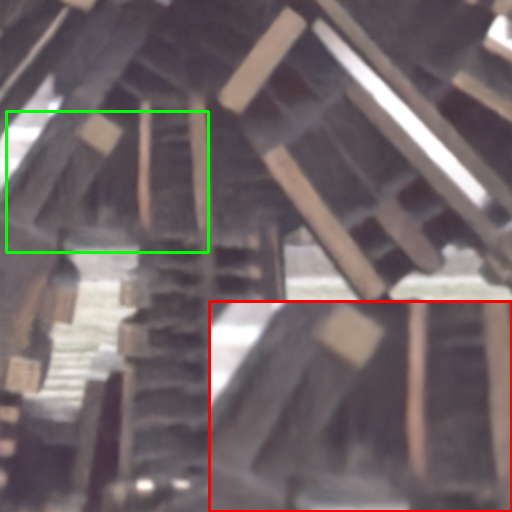}\\COLA-B (Ours)
\end{minipage}
\begin{minipage}[t]{0.12\linewidth}
\centering
\includegraphics[width=1\columnwidth]{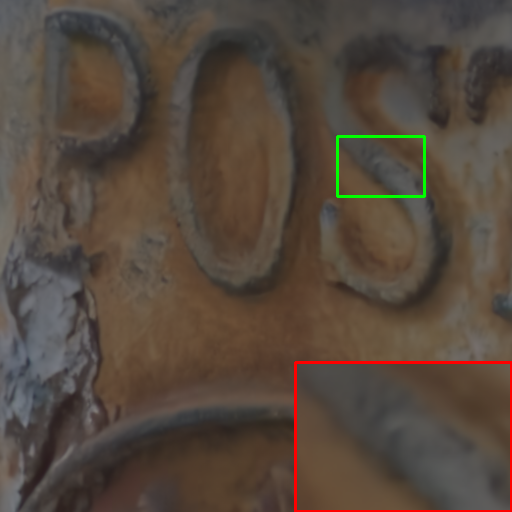}\\COLA-E (Ours)\\~\\
\includegraphics[width=1\columnwidth]{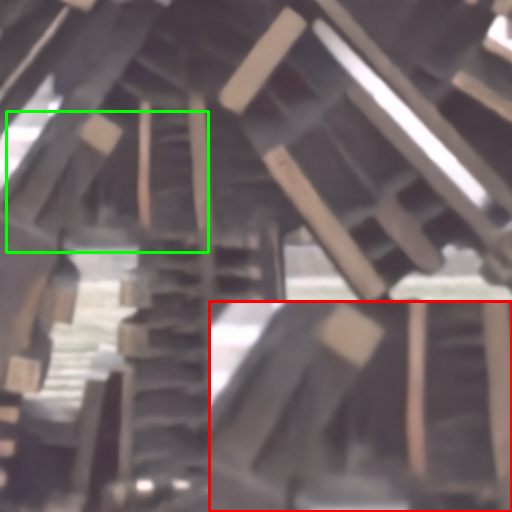}\\COLA-E (Ours)
\end{minipage}
\centering
\caption{Visual comparison of real image denoising application of various methods. These noisy images are corrupted by realistic noise from DND \cite{dnd} dataset.}
\label{im_rdn} 
\end{figure*}

\subsubsection{\textbf{Real Image Denoising}}
To further prove the merits of our proposed COLA-Net, we apply it to the more challenging task of real image denoising. Different from synthetic image denoising, in this case, images are corrupted by realistic noise, which can be well explained by a Poisson-Gaussian distribution and can be further approximated with a heteroscedastic Gaussian distribution defined as below:
\begin{equation}
\begin{cases}
 & n(L)\sim \mathcal{N}(0,\sigma^{2}(L)) \\ 
 & \sigma^{2}(L)=L\cdot \sigma_{s}^{2} + \sigma_{c}^{2},
\end{cases}
\label{eq_rn}
\end{equation}
where $L$ is the irradiance intensity of raw pixels, $\sigma_{s}^{2}$ and $\sigma_{c}^{2}$ refer to the spatially variant noise variance and stationary noise variance, respectively. Since it is generally expensive to acquire hundreds of noisy images with realistic noise and their corresponding high-quality labels, existing real noise datasets are relatively small which can not be used to train neural networks adequately. Thus, this approximate noise model is widely used to perform data argumentation in real image denoising tasks.

\begin{table}[t]
\caption{The quantitative results on the DND benchmark.}
\centering
\begin{tabular}{|c|c|c|c|c|}
\hline
\multirow{2}*{Algorithm}&\multirow{2}*{Parameters}&\multirow{2}*{Blind/Non-blind}&\multicolumn{2}{c|}{sRGB}\\
\cline{4-5}
&&& PSNR & SSIM \\
\hline
\hline
BM3D \cite{BM3D}&-& Non-blind & 34.51& 0.851 \\
CDnCNN \cite{dncnn}&0.67M& Blind & 32.43 & 0.790 \\
CFFDNet \cite{ffdnet}&0.85M &Non-blind & 37.61 & 0.914\\
TWSC \cite{twsc}&- & Blind& 37.94 & 0.940 \\
CBDNet \cite{cbdnet}&4.36M &Blind & 38.06 & 0.942\\
VDNet \cite{vdnet} & 7.82M & Blind & 39.38 & 0.952\\
CycleISP \cite{cycleisp} & 2.60M & Blind & 39.56 & 0.956\\
AINDNet \cite{aindnet} & 13.76M &Blind & \underline{39.77} & \textbf{0.959} \\
MIRNet \cite{mirnet} & 31.79M & Blind &\textbf{39.88} & \underline{0.956} \\
\hline
\hline
COLA-B&1.18M& Blind & 39.07 & 0.949\\
\hline
COLA-E&1.92M& Blind & 39.64 & 0.954\\
\hline
\end{tabular}
\vspace{-5pt}
\label{tb_rdn}
\end{table}

\begin{table*}[h]
\caption{Ablation study of different components in COLA-Net. PSNR values are evaluated on Urban100 ($\sigma = 25$).}
\centering
\begin{tabular}{|c|c|c|c|c|c|c|c|c|c|}
\hline
  Case Index & 0 (default)  & 1  & 2 & 3 & 4 & 5 & 6 & 7 & 8\\
\hline
\hline
Patch-wise Non-local Attention & $\checkmark$ & $\times$  & $\times$ & $\checkmark$ & $\times$ &$\times$ & $\checkmark$ & $\checkmark$ & $\checkmark$ \\
Local Attention & $\checkmark$ &$\times$ & $\checkmark$ & $\times$ & $\checkmark$ & $\times$ & $\checkmark$ & $\checkmark$ & $\checkmark$\\
Pixel-wise Non-local Attention \cite{nonlocalnet} & $\times$&$\checkmark$ & $\times$ & $\times$ & $\checkmark$ & $\times$ & $\times$ & $\times$ & $\times$\\
Number of CAB& 4 & 4 & 4 & 4&4 & 4 & 3 & 2 & 1\\
\hline
\hline
PSNR (dB) & 31.17& 30.70 & 30.18 & 31.04 & 30.75 & 29.79 & 31.12 & 31.01 & 30.68\\
\hline
Parameters &1.10M&0.98M &1.00M &1.06M&1.02M &0.88M &0.97M&0.70M&0.42M\\
\hline
\end{tabular}
\label{tb_as}
\end{table*}

During training, the heteroscedastic Gaussian distribution (Eq.~\ref{eq_rn}) is used to generate the synthetic training samples. The commonly used DND \cite{dnd} dataset is used for evaluation, which contains 50 images corrupted by realistic noise, and their high-quality labels are not available. We get the quantitative results from the benchmark website: \textbf{https://noise.visinf.tu-darmstadt.de/}, and also release the official evaluation result of our method on this website. The objective comparison is shown in Table~\ref{tb_rdn}. It is clear to see that the basic COLA-Net (COLA-B) achieves attractive performance with much fewer parameters, and the enhanced COLA-Net (COLA-E) outperforms some very recent competitive methods (\textit{e.g.,} CBDNet \cite{cbdnet}, VDNet \cite{vdnet} and CycleISP \cite{cycleisp}) while maintaining an attractive computational complexity. Even compared with existing top-performing methods \cite{aindnet,mirnet}, our method can achieve comparable performance with much fewer parameters (only about $\frac{1}{8}$ of \cite{aindnet} and $\frac{1}{17}$ of \cite{mirnet}). Fig.~\ref{im_rdn} further provides a visual comparison of different methods on two samples from DND dataset. Compared with other methods, COLA-Net is able to produce higher visual quality with much sharper image contents and fewer noisy signals, especially in the boundary areas, which benefits from the proposed collaborative attention mechanism.

\subsection{Ablation Study}
In this subsection, we show the ablation study in Table \ref{tb_as} to investigate the effect of different components in COLA-Net. Note that Case 0 denotes our basic COLA-Net with the default setting. In the ablation study, we compare our method with several baseline models: $(Case 1)$ We replace the dual-branch fusion modules with non-local neural networks \cite{nonlocalnet}. $(Case 2)$ We replace our patch-wise non-local attention submodule with non-local neural networks \cite{nonlocalnet}. $(Case 3)$ We remove all patch-wise non-local attention submodules from our COLA-Net. $(Case 4)$ We remove all local attention submodules. $(Case 5)$ We remove all dual-branch fusion modules. The detailed analysis of the ablation study is presented below.     


\begin{itemize}

\item \textbf{Non-local attention.} Compared with Case 0, Case 2 removes the non-local attention submodule. The obvious performance decrease in Case 2 indicates the positive effect of our proposed non-local attention operation. In Case 3, we only apply non-local attention submodule, and in Case 5, we remove both local and non-local attention submodules. The performance comparison between Case 3 and Case 5 demonstrates the necessity of the non-local attention submodule.

\item \textbf{Local Attention.} To verify the effect of local attention submodules, we also provide two groups of comparison. From Case 0 and Case 3, one can see that local attention submodule brings 0.13 dB PSNR gains together with patch-wise non-local attention submodule. From Case 2 and Case 5, the local attention submodule contributes 0.39 dB PSNR gains.   

\item \textbf{Patch-wise Non-local Attention.} To verify the effect of our proposed patch-wise non-local attention model, we make a comparison between Case 0 and Case 4, as well as between Case 3 and case 1. The 0.42 dB PSNR gains by Case 0 over Case 4 and 0.34dB gains by Case 3 over Case 1 clearly show the superiority of our proposed patch-wise non-local attention operation over existing pixel-wise non-local attention operation \cite{nonlocalnet}.

\item \textbf{Collaborative Attention.} 
To verify the necessity of collaborative attention, we compare it with only local attention and only non-local attention separately. The 0.99 dB gains by Case 0 over Case 2 and the 0.13 dB gains by Case 0 over Case 3 fully demonstrate the superiority of our proposed collaborative attention block.

\item \textbf{Number of CAB.}
To study the effect of the number of CAB, we make a comparison among Cases 0, 6, 7, and 8, which clearly shows that the performance increases with the number of CAB. By making a trade-off between performance and complexity, the default number of CAB is set to be 4. It is also worth emphasizing that our proposed COLA-Net with only one CAB, that is, Case 8, still achieves better performance than ADNet \cite{dilated2} and N3Net \cite{n3net}, despite having fewer parameters (425K vs. 560K and 720K) and fewer layers (12 vs. 17 and 24).

\end{itemize}

\section{Conclusion}
In this paper, a novel \textbf{COL}laborative \textbf{A}ttention \textbf{Net}work, dubbed \textbf{COLA-Net} is proposed by incorporating both local and non-local attention mechanisms into deep network for image restoration tasks. COLA-Net is the first attempt to adaptively combine local and non-local operations and to enable restoring complex textures and repetitive details distinguishingly. An effective and robust patch-wise non-local attention model is also developed to establish a more reliable long-range dependence during image restoration. 
Extensive experiments on three typical image restoration tasks, i.e., synthetic image denoising, real image denoising and compression artifact reduction, show that the proposed COLA-Net achieves state-of-the-art results, while maintaining an attractive computational complexity.

\ifCLASSOPTIONcaptionsoff
  \newpage
\fi

{
\bibliographystyle{IEEEtran}
\bibliography{egbib}
}

\end{document}